\documentclass[letterpaper]{article} % DO NOT CHANGE THIS
\usepackage{aaai2026}  % DO NOT CHANGE THIS
\usepackage{times}  % DO NOT CHANGE THIS
\usepackage{helvet}  % DO NOT CHANGE THIS
\usepackage{courier}  % DO NOT CHANGE THIS
\usepackage[hyphens]{url}  % DO NOT CHANGE THIS
\usepackage{graphicx} % DO NOT CHANGE THIS
\usepackage{natbib}  % DO NOT CHANGE THIS AND DO NOT ADD ANY OPTIONS TO IT
\usepackage{caption} % DO NOT CHANGE THIS AND DO NOT ADD ANY OPTIONS TO IT
\usepackage{algorithm}

\usepackage{algorithm}
\usepackage{algpseudocode}

\usepackage{amsmath}
\usepackage{amssymb}
\usepackage{amsthm, bm}
\usepackage{booktabs}
\usepackage{multirow}
\usepackage{newfloat}
\usepackage{listings}
\DeclareCaptionStyle{ruled}{labelfont=normalfont,labelsep=colon,strut=off} % DO NOT CHANGE THIS
\floatstyle{ruled}
\newfloat{listing}{tb}{lst}{}
\floatname{listing}{Listing}
\title{MARS: \underline{M}ulti-Agent \underline{A}daptive \underline{R}easoning with \underline{S}ocratic Guidance\\
for Automated Prompt Optimization}

\author{
    Jian Zhang\equalcontrib\textsuperscript{\rm 1,2},
    Zhangqi Wang\equalcontrib\textsuperscript{\rm 1,3},
    Haiping Zhu\textsuperscript{\rm 1,2}\thanks{Corresponding Author},
    Kangda Cheng\textsuperscript{\rm 4},\\
    Kai He\textsuperscript{\rm 5},
    Bo Li\textsuperscript{\rm 1,3},
    Qika Lin\textsuperscript{\rm 5}\footnotemark[2],
    Jun Liu\textsuperscript{\rm 1,3},
    Erik Cambria\textsuperscript{\rm 6}
}
\affiliations{
    \textsuperscript{\rm 1}School of Computer Science and Technology, Xi’an Jiaotong University, China\\
    \textsuperscript{\rm 2}MOE KLINNS Lab, Xi’an Jiaotong University, China\\
    \textsuperscript{\rm 3}Shaanxi Province Key Laboratory of Big Data Knowledge Engineering, Xi’an Jiaotong University, China\\
    \textsuperscript{\rm 4}School of Electronics and Information Engineering, Harbin Institute of Technology, Harbin, China\\
    \textsuperscript{\rm 5}Saw Swee Hock School of Public Health, National University of Singapore, Singapore\\
    \textsuperscript{\rm 6}College of Computing and Data Science, Nanyang Technological University, Singapore\\
    zhangjian062422@stu.xjtu.edu.cn, 
    zhuhaiping@xjtu.edu.cn
}
\usepackage{bibentry}
\begin{document}

\maketitle

\begin{abstract}
Large language models (LLMs) typically operate in a question-answering paradigm, where the quality of the input prompt critically affects the response. Automated Prompt Optimization (APO) aims to overcome the cognitive biases of manually crafted prompts and explore a broader prompt design space. However, existing APO methods often suffer from rigid template structures and inefficient exploration in the prompt space.
To this end, we propose a \underline{\textbf{M}}ulti-Agent \underline{\textbf{A}}daptive \underline{\textbf{R}}easoning with \underline{\textbf{S}}ocratic guidance framework (\textbf{MARS}) for APO. MARS consists of five complementary agents and formulates the optimization process as a Partially Observable Markov Decision Process (POMDP), enabling adaptive prompt refinement through explicit state modeling and interactive feedback.
Specifically, a \textit{Planner} agent generates flexible optimization trajectories, a \textit{Teacher}-\textit{Critic}-\textit{Student} triad engages in Socratic-style dialogue to iteratively optimize the prompt based on pseudo-gradient signals in the text space, and a \textit{Target} agent executes the prompt in downstream tasks to provide performance feedback. MARS integrates reasoning, feedback, and state transition into a unified hidden-state evolution process, improving both the effectiveness and interpretability of optimization.
Extensive experiments on multiple datasets demonstrate that MARS outperforms existing APO methods in terms of optimization performance, search efficiency, and interpretability.
\end{abstract}

\begin{links}
    \link{Code}{https://github.com/exoskeletonzj/MARS}
    % \link{Datasets}{https://aaai.org/example/datasets}
    % \link{Extended version}{https://aaai.org/example/extended-version}
\end{links}

\section{Introduction}

\begin{figure}[t]
 \includegraphics[width=\columnwidth]{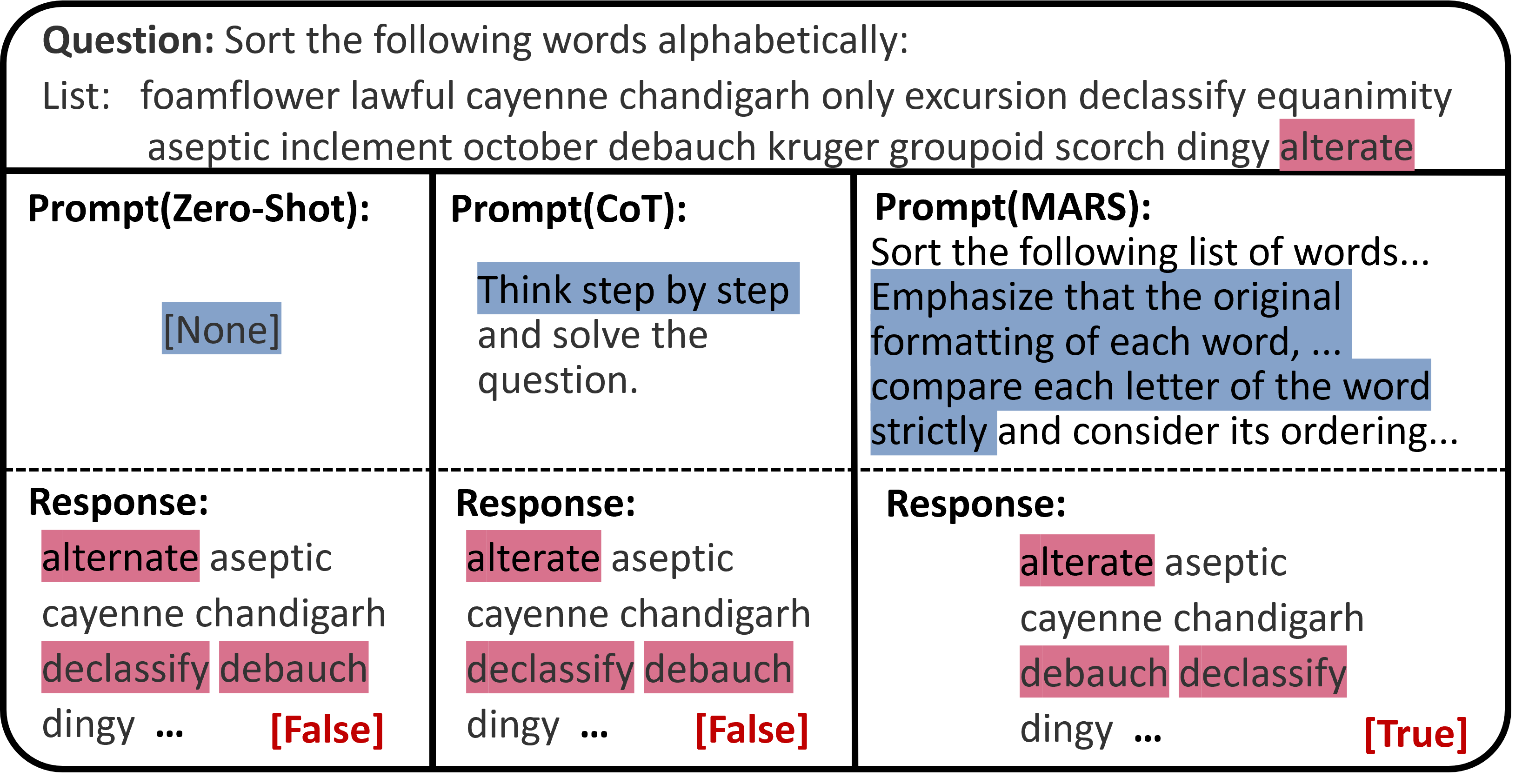}
 \caption{
 % For the word sorting task, three different prompts along with their corresponding response examples.
 Three different prompts along with their corresponding responses for the word sorting task.
 }
 \label{fig_example}
\end{figure}

Large language models (LLMs) such as GPT-4~\citep{achiam2023gpt} and Deepseek-R1~\citep{guo2025deepseek} provide robust support for thousands of natural language processing tasks~\citep{ijcai2025p772}. By providing a natural language prompt that includes instructions and a task description, LLMs can quickly adapt and respond~\citep{lin2025self}~\citep{shentie2025}~\citep{yan2025mur}. Consequently, the quality of the prompt is of critical importance, leading to wide interest in Automated Prompt Optimization (APO)~\citep{pryzant2023automatic}.
As shown in Figure~\ref{fig_example}, we provide LLMs with three different inputs for the word sorting task: a zero-shot prompt, a Chain of Thought (CoT) prompt, and our optimized prompt.
% They yield three distinct results. 
The responses are produced in a markedly distinct way.
Specifically, the zero-shot prompt incorrectly identifies the \emph{alterate} as the more common word \emph{alternate}. However, the task requires faithfully preserving the given sequence of words rather than correcting them. With the CoT prompt, the sorting remains incorrect because the LLM does not fully grasp the sorting task and the word sequence. In contrast, our optimized prompt produces the correct answer. This is because our prompt includes specific requirements, such as maintaining the original letter casing and specifying the sorting method.

\begin{figure*}[t]
	\large
	\centering
	\includegraphics[scale=0.4]{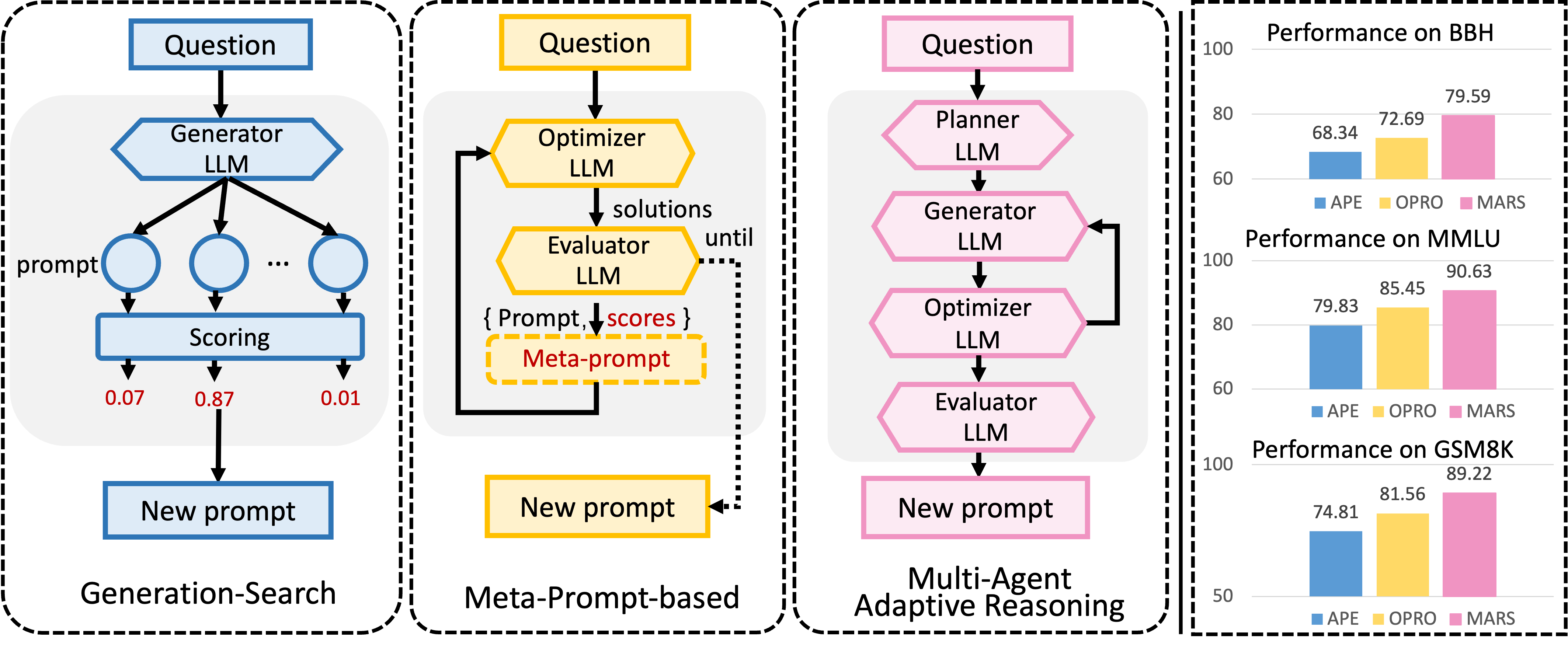}
	\caption{
Comparison of APO strategies. Generation-search and meta-prompt. Multi-Agent Adaptive Reasoning enables dynamic, collaborative reasoning. Right: With GPT-4o, MARS outperforms all baselines on three benchmarks.}
	\label{compare}
\end{figure*}

Thus, it is evident that APO can lead to improved performance in downstream tasks. 
As shown in Figure~\ref{compare}, recent studies~\citep{zhoularge, xureprompting, wang2023promptagent} have explored prompt optimization by generating multiple candidates combined with diverse search strategies, while others~\citep{Yang0LLLZC24, ye2023prompt} focus on designing sophisticated \emph{meta-prompts} to guide optimization. 
Despite these advances, two key issues remain: the limited flexibility of fixed prompt templates, and the inefficiency of prompt space exploration.

The first issue is the \textbf{limited flexibility of fixed templates}. Prior works~\citep{Yang0LLLZC24, ye2023prompt} often rely on \emph{meta-prompts}, which are predefined optimization templates that cannot be dynamically adapted to different tasks. Unlike domains such as event extraction~\citep{zhang2024semantic} or text-to-symbol generation~\citep{xu2023symbol}, where fixed templates suffice due to the task's structural regularity, APO requires more adaptability. Rigid templates may introduce biases or fail to capture task-specific nuances, resulting in suboptimal performance when applied to diverse or complex scenarios.

The second issue is the \textbf{inefficiency of prompt space exploration}. Several approaches~\citep{zhoularge, xureprompting, wang2023promptagent} adopt a \emph{generation-search} strategy, where a set of candidate prompts is first generated and then refined using local search techniques. However, this approach typically performs only local exploration around the initial candidates, without sufficiently covering the broader prompt space. As a result, the optimization may converge prematurely or overlook better-performing prompts, limiting overall effectiveness.

To this end, we propose a \textbf{M}ulti-Agent \textbf{A}daptive \textbf{R}easoning with \textbf{S}ocratic guidance framework (\textbf{MARS}) for APO. MARS consists of five complementary agents and formulates the optimization process as a Partially Observable Markov Decision Process (POMDP), enabling adaptive prompt refinement through explicit state modeling and interactive feedback.
 Functionally, to address the first challenge, MARS introduces a \textit{Planner} agent that generates task-specific optimization trajectories, allowing prompts to be flexibly adapted to diverse task requirements. To tackle the second challenge, MARS employs a Socratic-style \textit{Teacher}-\textit{Critic}-\textit{Student} dialogue mechanism, which iteratively guides prompt refinement. This module enables effective exploration of the prompt space by simulating a gradient-inspired optimization process, while also promoting interpretability.
The overall process is modeled as a POMDP, where the hidden state represents the latent reasoning state of the \textit{Student} agent. Through multi-agent interactions and performance feedback from a \textit{Target} agent, MARS approximates a pseudo-gradient trajectory in the discrete prompt space, progressively refining the prompt toward an optimal solution.
%We conduct extensive experiments on both general-purpose and domain-specific benchmarks to validate the effectiveness of MARS. Further analysis demonstrates the interpretability of its optimization trajectories and the adaptability of its agent-based architecture. 
Our contributions are three-fold:

$\bullet$\;This work is the first to introduce a multi-agent architecture with POMDP modeling for APO. It proposes MARS, which enables hidden-state reasoning and adaptive planning through agent collaboration.

$\bullet$\;A \textit{Teacher}-\textit{Critic}-\textit{Student} Socratic dialogue mechanism is designed to enable interpretable, iterative prompt refinement via a gradient-inspired optimization trajectory.

$\bullet$\;We demonstrate the effectiveness and generalizability of MARS through extensive experiments on both general and domain-specific benchmarks, and validate the interpretability of its optimization process.

\begin{figure*}[t]
	\large
	\centering
	\includegraphics[scale=0.5]{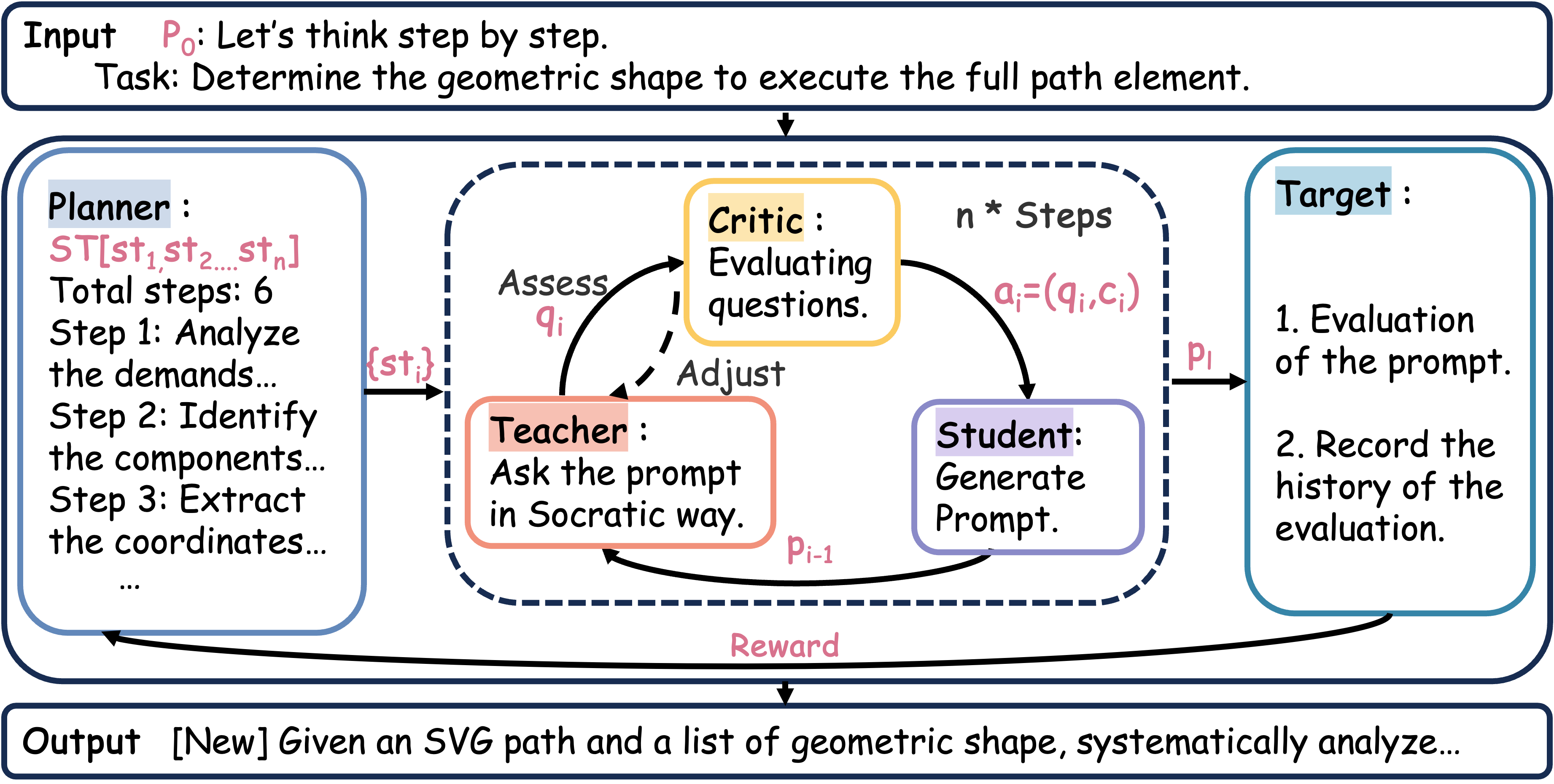}
	\caption{
  % The MARS model architecture diagram.
  The overall architecture of the MARS model.
  It consists of five LLM agents. The \textit{Planner} agent that autonomously generates task-specific optimization trajectories, and a \textit{Teacher}-\textit{Critic}-\textit{Student} Socratic dialogue mechanism that iteratively refines prompts, with the evaluation and iterative refinement process guided by feedback from the \textit{Target} agent.}
	\label{fig_model}
\end{figure*}

\section{Methodology}

MARS comprises two main modules: (i) a high-level \textit{Planner} that generates task-specific optimization trajectories, and (ii) a \textit{Teacher}-\textit{Critic}-\textit{Student} triad that performs Socratic-style iterative refinement.
The overall architecture is shown in Figure~\ref{fig_model}, with the complete workflow detailed in Algorithm~\ref{alg:training}. This section introduces: (1) the APO task and its POMDP formulation, (2) the \textit{Planner} design, (3) the gradient-inspired Socratic dialogue mechanism, and (4) the evaluation-feedback loop via the \textit{Target} agent.

\subsection{Task Formulation and POMDP Modeling} \label{sec:taskdefinition}

Given a task-specific \textit{Target} model \(\mathcal{M}_{\text{tar}}\), the goal of APO is to iteratively refine an initial prompt \(p_0\) to an optimal version \(p^*\) that maximizes performance on a downstream dataset \(D = \{(x, y)\}\). A training subset \(D_{\text{train}} \subset D\) guides the optimization, while \(D_{\text{test}}\) is used for evaluation. The objective can be formalized as:
\begin{equation}
p^* = \arg\max_{p} \sum_{(x,y)\in D_{\text{test}}} f\left(\pi_{\text{tar}}(x; p), y\right),
\end{equation}
where \(\pi_{\text{tar}}(x; p)\) denotes the model output conditioned on \(x\) and prompt \(p\), and \(f\) is a task-specific metric (e.g., accuracy, BLEU).

To capture the sequential, partially observable nature of the optimization, we model APO as a Partially Observable Markov Decision Process (POMDP) defined by:
\[
\langle \mathcal{S}, \mathcal{A}, \mathcal{T}, \mathcal{R}, \mathcal{O} \rangle,
\]
where:
- \(\mathcal{S}\): latent state space, representing the internal reasoning state of the \textit{Student} agent;
- \(\mathcal{A}\): action space, comprising instructional signals (e.g., questions, critiques) from \textit{Teacher} and \textit{Critic};
- \(\mathcal{T}\!: \mathcal{S} \times \mathcal{A} \rightarrow \mathcal{S}\): transition dynamics, updating student states;
- \(\mathcal{O}\!: \mathcal{S} \rightarrow \mathcal{P}\): observation function, mapping hidden states to prompts;
- \(\mathcal{R}(s, a) = f(\pi_{\text{tar}}(x; \mathcal{O}(s)), y)\): reward function, based on performance of the generated prompt.

This formulation allows MARS to perform gradient-inspired prompt refinement in a partially observable, discrete text space. Via iterative multi-agent reasoning and feedback, the system progressively transitions from \(p_0\) to \(p^*\).

\subsection{Optimization Trajectory Planning} \label{sec:plan}

As illustrated in Figure~\ref{fig_model}, MARS begins with a \textit{Planner} agent that initiates the prompt optimization process.

\noindent\textbf{\textit{Planner}}. Given task goal \(g\), input \(x \in D_{\text{train}}\), and initial prompt \(p_0\), the \textit{Planner} decomposes the optimization into a sequence of sub-goals:
\begin{equation}
\mathbf{ST} = [st_1, st_2, \dots, st_n] = \pi_{\text{plan}}(g, x, p_0),
\end{equation}
where \(\pi_{\text{plan}}\) is the planning policy that adaptively generates an optimization trajectory.

To formalize \(\pi_{\text{plan}}\), we introduce a latent planning variable \(z\in\mathcal{Z}\), and model trajectory generation as:
\begin{equation}
\pi_{\text{plan}}(g, x, p_0) = \arg\max_{\mathbf{ST}} \, \mathbb{E}_{z \sim q(z|g, x)} \left[ \log P(\mathbf{ST} \mid z, p_0) \right],
\end{equation}
where \(q(z|g, x)\) captures task semantics, and \(P(\mathbf{ST} \mid z, p_0)\) models the trajectory conditioned on latent intent and initial prompt.
This hierarchical formulation induces structured plans over latent space \(\mathcal{S}\), guiding local agent decisions under global coherence and improving adaptability over static templates.

\subsection{Socratic Prompt Refinement as Joint Policy Optimization} \label{sec:interaction}

Prompt refinement in discrete language space presents unique challenges due to its non-differentiability, high variance, and semantic ambiguity. To address these issues, MARS employs a structured Socratic dialogue mechanism involving three collaborative agents—\textit{Teacher} (\(\pi_{\text{t}}\)), \textit{Critic} (\(\pi_{\text{c}}\)), and \textit{Student} (\(\pi_{\text{s}}\))—each fulfilling a complementary role in exploring and improving prompts through guided interaction. This framework transforms prompt optimization into an interpretable, policy-driven reasoning process grounded in pedagogical principles.

At each refinement step \(i\), given a sub-goal \(st_i \in \mathbf{ST}\), the \textit{Teacher} proposes a Socratic-style question \(q_i\) to stimulate reasoning, based on the prior prompt \(p_{i-1}\). The \textit{Critic} then assesses its clarity, relevance, and coherence, producing feedback \(c_i\) to revise or validate the proposed direction. Finally, the \textit{Student} responds by updating its internal state and generating a new prompt \(p_i\). This process is formalized as:
\begin{equation}
\begin{aligned}
q_i &= \pi_{\text{t}}(st_i, p_{i-1}), \\
c_i &= \pi_{\text{c}}(q_i), \\
p_i &= \pi_{\text{s}}((q_i, c_i), p_{i-1}), \\
s_i &\sim \mathcal{T}(s_{i-1}, (q_i, c_i)), \quad o_i = p_i.
\end{aligned}
\end{equation}
Each agent performs a partial update to the joint optimization process: \textit{Teacher} drives semantic direction, \textit{Critic} provides quality control, and \textit{Student} synthesizes the final output.

\paragraph{Context-Aware Interaction.}
To improve reasoning consistency and avoid step-wise myopia, each agent conditions not only on the current sub-goal and prompt, but also on the dialogue history \(\mathcal{H}_{<i} = \{(q_j, c_j, p_j)\}_{j=1}^{i-1}\). The full context-aware behavior is given by:
\begin{equation}
\begin{aligned}
q_i &= \pi_{\text{t}}(st_i, p_{i-1}, \mathcal{H}_{<i}), \\
c_i &= \pi_{\text{c}}(q_i, \mathcal{H}_{<i}), \\
p_i &= \pi_{\text{s}}((q_i, c_i), p_{i-1}, \mathcal{H}_{<i}).
\end{aligned}
\end{equation}
By attending to prior reasoning steps, the system forms coherent, memory-informed trajectories across iterations.

\paragraph{Joint Optimization Objective.}
We define the multi-agent policy as \(\Pi = \{\pi_{\text{t}}, \pi_{\text{c}}, \pi_{\text{s}}\}\), and optimize it jointly to maximize task performance while ensuring interpretability and alignment with sub-goals:
\begin{equation}
\max_{\Pi} \, \mathbb{E}_{(x, y) \sim D} \left[ \mathcal{R}(\Pi) - \lambda \sum_{i=1}^{n} \mathcal{L}_{\text{align}}((q_i, c_i), st_i) \right],
\end{equation}
where \(\mathcal{R}(\Pi)\) denotes the cumulative reward from the \textit{Target} agent, and \(\mathcal{L}_{\text{align}}\) penalizes semantic drift from intended optimization goals.

This tri-agent structure enables interpretable, step-wise refinement of prompts through structured reasoning and localized feedback, offering both flexibility and transparency in discrete prompt optimization.

\textbf{Proposition 1 (Socratic Policy Improvement Bound).}
Let \(\Pi = \{\pi_{\text{t}}, \pi_{\text{c}}, \pi_{\text{s}}\}\) denote the joint policy, and suppose the Socratic signal \(a_i = (q_i, c_i)\) induces expected advantage \(\bar{A}_i > 0\) over the prior state \(s_{i-1}\). Then, under bounded variance \(\sigma^2\), the cumulative improvement over \(n\) steps satisfies:

\begin{equation}
\mathbb{E}[\mathcal{R}(p_n)] - \mathcal{R}(p_0) \ge \sum_{i=1}^n \left( \bar{A}_i - \frac{\sigma^2}{2\lambda} \right),
\end{equation}

where \(\lambda\) is the local Lipschitz constant of the reward surface.

\textit{This provides a lower bound on improvement, formally linking guidance signal quality to reward trajectory. }

 Derivation is in Appendix A.1.

\begin{figure}[t]
  \large
  \centering
 \includegraphics[width=\columnwidth]{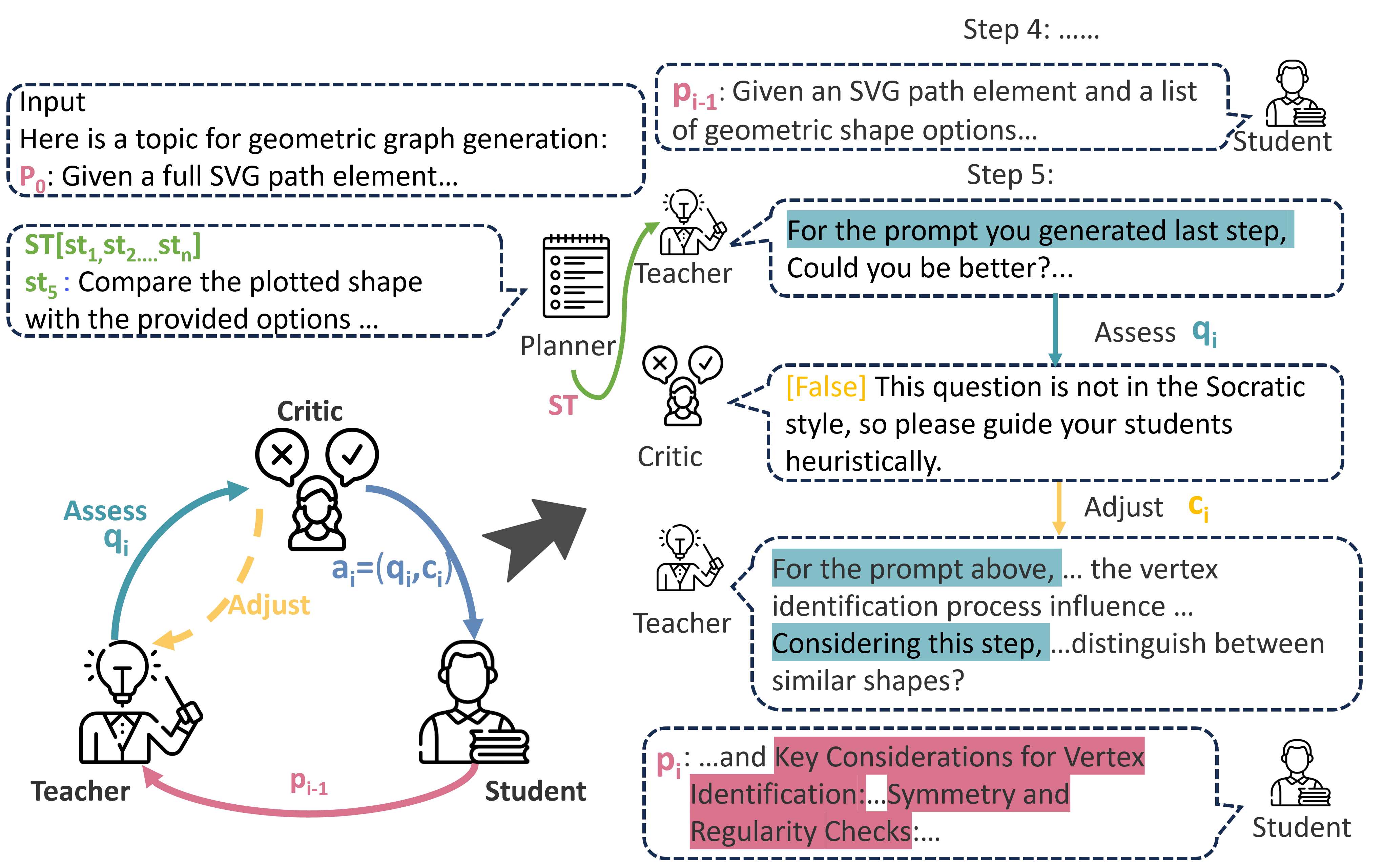}
 % \vspace{-0.6cm}
 \caption{A specific illustration of a \textit{Teacher}-\textit{Critic}-\textit{Student} Socratic guidance dialogue pattern. The case shows the fifth step optimization iteration.}
 \label{fig_example1}
 % \vspace{-0.4cm}
\end{figure}

\begin{algorithm}[t]
\footnotesize
\caption{MARS Optimization Procedure}
\label{alg:training}
\begin{algorithmic}[1]
\State \textbf{Input:} Dataset $\mathcal{D}$, initial prompt $p_0$, threshold $\delta$, max iterations $I$
\State \textbf{Output:} Optimized prompt $p^*$
\State \textbf{\textit{Planner}}: Generate sub-goal trajectory $\mathbf{ST} = \{st_1, \dots, st_n\}$
\State Initialize $p_0^{(1)} \gets p_0$, $\mathcal{R}^{(0)} \gets 0$
\For{iteration $t = 1$ to $I$}
  \For{step $i = 1$ to $n$} \hfill\textit{// Generate question}
  % \Comment{\textit{Socratic refinement}}
    \State \textbf{\textit{Teacher}} generates question $q_i \gets \pi_{\text{t}}(st_i, p_{i-1}^{(t)})$
    \Repeat
      \State \textbf{\textit{Critic}} evaluates $q_i$ \& returns feedback $c_i \gets \pi_{\text{c}}(q_i)$
      \State \textbf{\textit{Teacher}} revises $q_i$ if $c_i$ is unsatisfactory
    \Until{Socratic quality is satisfied}
    \State Set $a_i \gets (q_i, c_i)$
    \State \textbf{\textit{Student}} updates $p_i^{(t)} \gets \pi_{\text{s}}(a_i, p_{i-1}^{(t)})$
  \EndFor
  \State Let $p_\ell^{(t)} \gets p_n^{(t)}$ \hfill\textit{// Final prompt}
  % \Comment{\textit{Final prompt}}
  \State \textbf{\textit{Target}} evaluates reward 
  \State $\mathcal{R}^{(t)} = \sum_{(x, y) \in D_{\text{test}}} f(\pi_{\text{tar}}(x; p_\ell^{(t)}), y)$
  \If{$\mathcal{R}^{(t)} - \mathcal{R}^{(t-1)} < \delta$}
    \State \textbf{break} \hfill\textit{//Early stopping}
    % \Comment{\textit{Early stopping}}
  \EndIf
\EndFor
\State \textbf{return} $p^* \gets p_\ell^{(t)}$
\end{algorithmic}
\end{algorithm}

\subsection{Evaluation and Iteration} \label{sec:feedback}

Upon completing the Socratic refinement trajectory, the final prompt \(p_\ell = p_n\)—produced through successive dialogue transitions from latent state \(s_0\) to \(s_\ell\)—is evaluated by the \textit{Target} agent \(\pi_{\text{tar}}\) on the held-out test set \(D_{\text{test}}\). The evaluation provides an external signal to measure the effectiveness of the entire optimization trajectory:
\begin{equation}
\mathcal{R}^{(t)} = \sum_{(x, y) \in D_{\text{test}}} f\left( \pi_{\text{tar}}(x; p_\ell^{(t)}), y \right),
\end{equation}
where \(f(\cdot)\) is a task-specific scoring function (e.g., accuracy, BLEU, F1), and \(p_\ell^{(t)}\) denotes the final prompt obtained at iteration \(t\). This scalar reward serves as the global performance metric, closing the loop between prompt generation and task-level effectiveness.

\paragraph{Adaptive Termination.}
To ensure efficient convergence and prevent over-refinement, we adopt an adaptive early stopping criterion based on marginal reward improvement. The gain between two consecutive iterations is defined as:
\begin{equation}
\Delta \mathcal{R}^{(t)} = \mathcal{R}^{(t)} - \mathcal{R}^{(t-1)}.
\end{equation}
The refinement continues only if:
\begin{equation}
\Delta \mathcal{R}^{(t)} > \delta, \quad t < I,
\end{equation}
where \(\delta\) is a minimum improvement threshold, and \(I\) is the maximum number of allowed iterations.

This iterative control mechanism enforces a form of performance-aware policy halting under the POMDP framework. It ensures that MARS focuses on high-impact updates while avoiding excessive computation on marginally beneficial revisions. As a result, the system adaptively determines the optimal stopping point based on observable task performance.

\textbf{Proposition 2 (Monotonic Reward Stability).}  
Assume \(\mathcal{R}(p)\) is \(\lambda\)-Lipschitz and each step satisfies \(\|p_i - p_{i-1}\| \le \varepsilon\). Then the reward trajectory \(\{\mathcal{R}(p_i)\}\) satisfies:
\[
|\mathcal{R}(p_{i+1}) - \mathcal{R}(p_i)| \le \lambda \varepsilon.
\]
In particular, if \(\mathcal{R}(p_i) < \mathcal{R}(p_{i+1})\) for some \(i\), then improvement is bounded and monotonic.
\textit{This result guarantees bounded gain/loss and motivates early stopping under stable improvement.}

 Derivation is in Appendix A.2.

\begin{table*}[t]
% \centering
% \footnotesize
% \resizebox{\linewidth}{!}{
\setlength{\tabcolsep}{6pt} % tighter column spacing
\centering
% \small % use smaller font to avoid resizing
\begin{tabular}{ccccccc|cccccc|c}
  \toprule
    % ~ & Boolean & Disambigu- & Formal & Geometric & Ruin & 
    % Sports & College & College & Electrical & High school & Human & Marketing & avg \\
    % ~ & Expre. & ation QA & Fallacies. & Shapes & Names & Understand. & Biology & Medicine & Engineering & World History & Aging & Test & advance \\
    Models & \textbf{B.E} & \textbf{D.QA} & \textbf{F.F.} & \textbf{G.S.} & \textbf{R.N.} & 
    \textbf{S.U.} & \textbf{C.B.} & \textbf{C.M.} & \textbf{E.E.} & \textbf{W.H.} & \textbf{H.A.} & \textbf{M.T.} & \textbf{Avg.} \\
  \midrule
    Origin & 74.70 & 51.41 & 52.20 & 43.37 & 59.84 & 60.24 & 82.52 & 69.77 & 63.89 & 73.73 & 66.22 & 81.55 & 64.95 \\[2pt]
    CoT(ZS) & 80.32 & 54.22 & 59.44 & 47.39 & 67.07 & 67.87 & 83.91 & 73.25 & 74.31 & 76.27 & 68.47 & 84.98 & 69.79 \\[2pt]
    CoT(FS) & 81.93 & 57.43 & \underline{66.26} & 49.40 & 70.68 & 72.29 & 86.71 & 76.74 & \underline{79.17} & 78.81 & 72.07 & 90.99 & 73.54 \\[2pt]
    APE & 83.53 & 61.85 & 61.04 & 51.41 & 77.51 & 74.70 & 88.11 & 75.58 & 69.44 & 82.20 & 75.68 & 87.98 & 74.09 \\[2pt]
    ProTeGi & 83.93 & 63.86 & 62.65 & 52.21 & 80.32 & 76.71 & 90.91 & 78.49 & 73.61 & 84.75 & 77.48 & 90.56 & 76.29 \\[2pt]
    OPRO & 86.34 & \underline{66.67} & 63.45 & 53.81 & 83.13 & \underline{82.73} & \underline{93.70} & \underline{83.14} & 77.01 & 86.44 & 79.73 & 92.70 & \underline{79.07} \\[2pt]
    PE2 & \underline{87.95} & 65.46 & 63.86 & \underline{54.62} & \underline{84.34} & 75.90 & 93.01 & 81.40 & 76.39 & \underline{88.14} & \underline{81.08} & \underline{93.56} & 78.81 \\
  \midrule
    Ours & \textbf{93.17} & \textbf{71.89} & \textbf{74.70} & \textbf{59.44} & \textbf{90.36} & \textbf{87.95} & \textbf{97.90} & \textbf{86.05} & \textbf{84.03} & \textbf{93.22} & \textbf{85.59} & \textbf{97.00} & \textbf{85.11} \\
  % \midrule
  %   Max advance & ~ & ~ & ~ & ~ & ~ & ~ & ~ & ~ & ~ & ~ & ~ & ~ & 5.18 \\
  \bottomrule
\end{tabular}
\caption{In the performance comparison across 12 general tasks, we carefully select 6 representative subtasks from both BBH and MMLU, two commonly used evaluation benchmarks, to comprehensively assess MARS's performance in diverse general-task settings. The evaluation results of these subtasks indicate that MARS surpasses all existing baseline methods.}
\label{mainResults}
% \vspace{-0.4cm}
\end{table*}

\begin{table}[htbp]
\setlength{\tabcolsep}{6pt}
  \centering
  \small
  \begin{tabular}{cccc|c|c|c}
    \toprule
      \multirow{2}{*}{Models} & \multicolumn{3}{c|}{\textbf{Chinese}} & \textbf{Math} & \textbf{Law} 
      & \textbf{Avg.} \\
       & A.S. & U.R.P. & CL.M. & GSM. & L.A. 
       & \\ 
    \midrule
      Origin & 56.25 & 48.89 & 57.14 & 67.07 & 23.14 & 50.50 \\[2pt]
      CoT(ZS) & 59.38 & 53.33 & 61.90 & 70.26 & 30.57 & 55.09 \\[2pt]
      CoT(FS) & 65.63 & 57.78 & 66.67 & 77.54 & \underline{35.81} & 60.69 \\[2pt]
      APE & 65.63 & 62.22 & 71.43 & 74.81 & 29.69 & 60.76 \\[2pt]
      ProTeGi & 68.75 & 66.67 & 76.19 & 77.47 & 31.88 & 64.19 \\[2pt]
      OPRO & 71.88 & 73.33 & \underline{80.95} & 81.56 & 31.44 & 67.83 \\[2pt]
      PE2 & \underline{75.00} & \underline{77.78} & 76.19 & \underline{83.46} & 34.50 & \underline{69.39} \\
    \midrule
      MARS & \textbf{81.25} & \textbf{84.44} & \textbf{85.71} & \textbf{89.22} & \textbf{38.42} & \textbf{75.81} \\
    \bottomrule
  \end{tabular}  
  \caption{Performance comparison on three types of domain-specific tasks: Chinese, law, and mathematics. The Chinese domain consists of three datasets, while the law and mathematics domains each have one dataset.}
  \label{table2}
  % \vspace{-0.3cm}
\end{table}

\section{Experiments} \label{sec:experiments}

In this section, we present extensive experiments conducted on 12 general task datasets and 5 domain-specific datasets. We begin by introducing the datasets and hyperparameters, followed by the main experimental results. A detailed analysis of the efficiency of the proposed framework is also provided. 

 More detailed descriptions of the tasks and datasets can be found in Appendix B. 
 The abbreviations used for the tasks in Tables~\ref{mainResults} and~\ref{table2}, along with their full names and dataset descriptions, are provided in Appendix B. Baseline methods and additional experimental details are introduced in Appendix C.

\subsection{Experimental Settings}

\paragraph{Datasets.}  
We select a total of 17 datasets covering both general-purpose and domain-specific tasks. Specifically, we use 6 tasks from the Big-Bench Hard (BBH) suite~\citep{suzgun2022challenging} and 6 tasks from MMLU~\citep{wang2024mmlu} to represent general reasoning and knowledge-intensive benchmarks. For domain-specific evaluation, we include 3 Chinese subject-area tasks from C-Eval~\citep{huang2024c}, 1 legal reasoning task from LSAT-AR~\citep{zhong2023agieval}, and 1 arithmetic reasoning task from GSM8K~\citep{zhang2024careful}.

\paragraph{Hyperparameters and Evaluation Protocol.}  
We adopt \texttt{deepseek-v2.5-1210}~\citep{guo2025deepseek} as the primary backbone LLM for all APO tasks. The generation temperature is set to 0.6 to balance creativity and coherence. We configure the maximum number of optimization iterations as \( I = 10 \), with an early stopping threshold of \( \delta = 0.01 \) based on accuracy improvement. To enhance efficiency, each \textit{assess-adjust} cycle is limited to a single revision per step. Final evaluation is performed using accuracy, computed by comparing the model prediction \( y_{\text{pred}} \) with the ground truth label \( y \).

\subsection{Main Results} \label{3.2}

\paragraph{MARS enhances the average performance across diverse task types.} 
The experimental results in Table~\ref{mainResults} and Table~\ref{table2} present a comprehensive comparison between the prompts optimized by MARS and the baselines for the 12 tasks. As shown in Table~\ref{mainResults}, on general tasks, MARS outperforms the previous SOTA by 6.04\%, and exceeds the original prompt and CoT(ZS) by 20.16\% and 15.32\%, respectively. This indicates that the prompts optimized by MARS enable LLMs to better understand the task requirements, providing stronger instructions for tasks across different scenarios.
MARS surpasses existing APO methods, highlighting the limitations of both the \emph{generate-search} approach and the \emph{meta prompts} approach. These methods fail to fully grasp the deeper essence of the APO process, which constrains their optimization effectiveness. In contrast, MARS thoughtfully considers the prompt optimization pathways for different tasks and incorporates heuristic optimization strategies, making the prompt refinement process more efficient and precise.

\begin{figure}[t]
  \large
  \centering
 \includegraphics[width=\columnwidth]{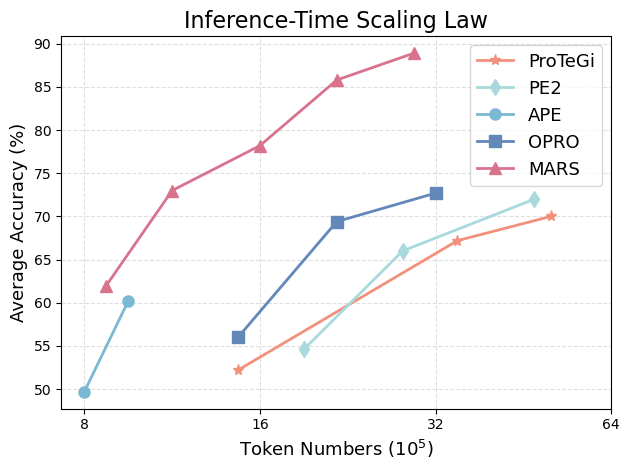}
 % \vspace{-1.2cm}
 \caption{Inference-time scaling law. The horizontal axis denotes the inference-time
computational cost, while the vertical axis represents
the average performances on all tasks.}
 \label{fig_example3}
 % \vspace{-0.3cm}
\end{figure}

\textbf{MARS achieves strong and consistent performance gains across domain-specific tasks, highlighting its effectiveness in knowledge-intensive reasoning.} Table~\ref{table2} presents the experimental results of MARS on domain-specific tasks, covering areas like Chinese, law, and mathematics, all of which require specialized knowledge and reasoning. In these tasks, MARS outperforms the previous SOTA methods to 6.42\%, further demonstrating its ability to better guide LLMs in domain-specific knowledge discovery and application. This not only lowers the barrier to utilizing LLMs but also enhances their generalization capability. Moreover, compared to the original prompt and CoT(ZS), MARS achieves improvements of 25.31\% and 20.72\%, respectively, underscoring its effectiveness and practicality in these specialized domains.

\subsection{Efficiency Analysis} \label{3.3}

\paragraph{MARS Consistently Achieves the Highest Computational Efficiency.}
The balance between resource consumption and performance improvement is a crucial analysis metric~\citep{yang2024harnessing}. 
As shown in Figure~\ref{fig_example3}, MARS consistently outperforms all baseline methods in terms of computational efficiency, as demonstrated by its superior inference-time scaling behavior across multiple APO tasks.

Notably, under the same number of output tokens, MARS achieves the highest performance across all evaluated tasks. Conversely, to reach comparable performance levels, baseline methods require more output tokens—indicating higher computational cost.

These results highlight MARS’s strong ability to balance performance and resource usage through its structured optimization strategy. By performing high-level task planning followed by step-wise Socratic refinement, MARS enables more efficient resource allocation, reduces unnecessary computation, and ensures both effectiveness and robustness throughout the APO process.

\begin{table}[t]
  \centering
  \setlength\tabcolsep{1pt} 
  \small % Reduce font size
  \begin{tabular}{l|ccccccc}
    \toprule
    \textbf{Variation} & \textbf{B.E.} & \textbf{D.QA} & \textbf{F.F.} & \textbf{G.S.} & \textbf{R.N.} & \textbf{S.U.} & \textbf{Avg.}\\
    \midrule
    MARS & 93.17 & 71.89 & 74.70 & 59.44 & 90.36 & 87.95 & 79.59\\
    \midrule
    ${w/o}_\text{Plan}$ & 86.35 & 65.86 & 68.67 & 54.21 & 82.33 & 79.52 & 72.82\\
    $\quad\Delta$ &  (-6.82) &  (-6.03) &  (-6.03) &  (-5.23) &  (-8.03) &  (-8.43) &  (-6.77) \\
    ${w/o}_\text{Soc}$ & 84.74 & 63.86 & 62.25 & 49.80 & 74.30 & 74.70 & 68.28\\
    $\quad\Delta$ &  (-8.43) &  (-8.03) &  (-12.45) &  (-9.64) &  (-16.06) &  (-13.25) &  (-11.31) \\
    ${w/o}_\text{Cri}$ & 89.16 & 68.27 & 72.28 & 56.22 & 86.34 & 83.94 & 76.04\\
    $\quad\Delta$ &  (-4.01) &  (-3.62) &  (-2.42) &  (-3.22) &  (-4.02) &  (-4.01) &  (-3.55) \\
    \bottomrule
  \end{tabular}
  \caption{Performance under different ablation settings are analyzed. We performed ablation experiments on the planner module ${w/o}_\text{Plan}$, the \textit{Teacher}-\textit{Critic}-\textit{Student} module ${w/o}_\text{Soc}$, and the \textit{Critic} Agent ${w/o}_\text{Cri}$ to evaluate the impact of removing these components. \textit{w/o} indicates the experiment was run without the specified module.}
  \label{ablation}
  % \vspace{-0.2cm}
\end{table}

\section{Supplementary Analysis}

To further validate the effectiveness of MARS, we conduct three additional analyses in this section: an ablation study to assess the contribution of each component, a convergence analysis to examine the optimization stability over iterations, and an investigation of the sensitivity to Other \textit{Target} LLMs.

 In addition, Appendix D reports MARS’s generalization performance when applied to GPT-4o and different initial prompt \(p_0\). Appendix E analyzes the effect of sample size on performance. Appendix F provides the internal prompts used by each agent to clarify their roles. Appendix G presents the full multi-agent interaction process on a representative APO task. Appendix H offers the optimized prompts for all 17 tasks to facilitate reproducibility.

\subsection{Ablation Study} \label{4.1}

\textbf{The Socratic dialogue mechanism plays the most critical role in MARS, as shown by the largest performance drop upon its removal.} Table~\ref{ablation} presents the impact of removing three key components: the \textit{Planner} agent, the \textit{Teacher-Critic-Student} Socratic module, and the \textit{Critic} agent. Removing the Socratic module leads to the most substantial degradation, as the system loses its iterative refinement capability and sends unprocessed sub-goals directly to the \textit{Target} agent, resulting in poor optimization quality. Eliminating the \textit{Planner} also causes a notable drop, since the Socratic dialogue lacks structured guidance without its sub-goal trajectory. Finally, while the \textit{Critic} contributes less overall, its feedback loop with the \textit{Teacher} improves prompt quality; removing it leads to a 3.55\% performance loss, as shown in Table~\ref{ablation}.

\subsection{Converagence Analysis} \label{4.2}

\begin{figure}[htp]
  \large
  \centering
 \includegraphics[width=\columnwidth]{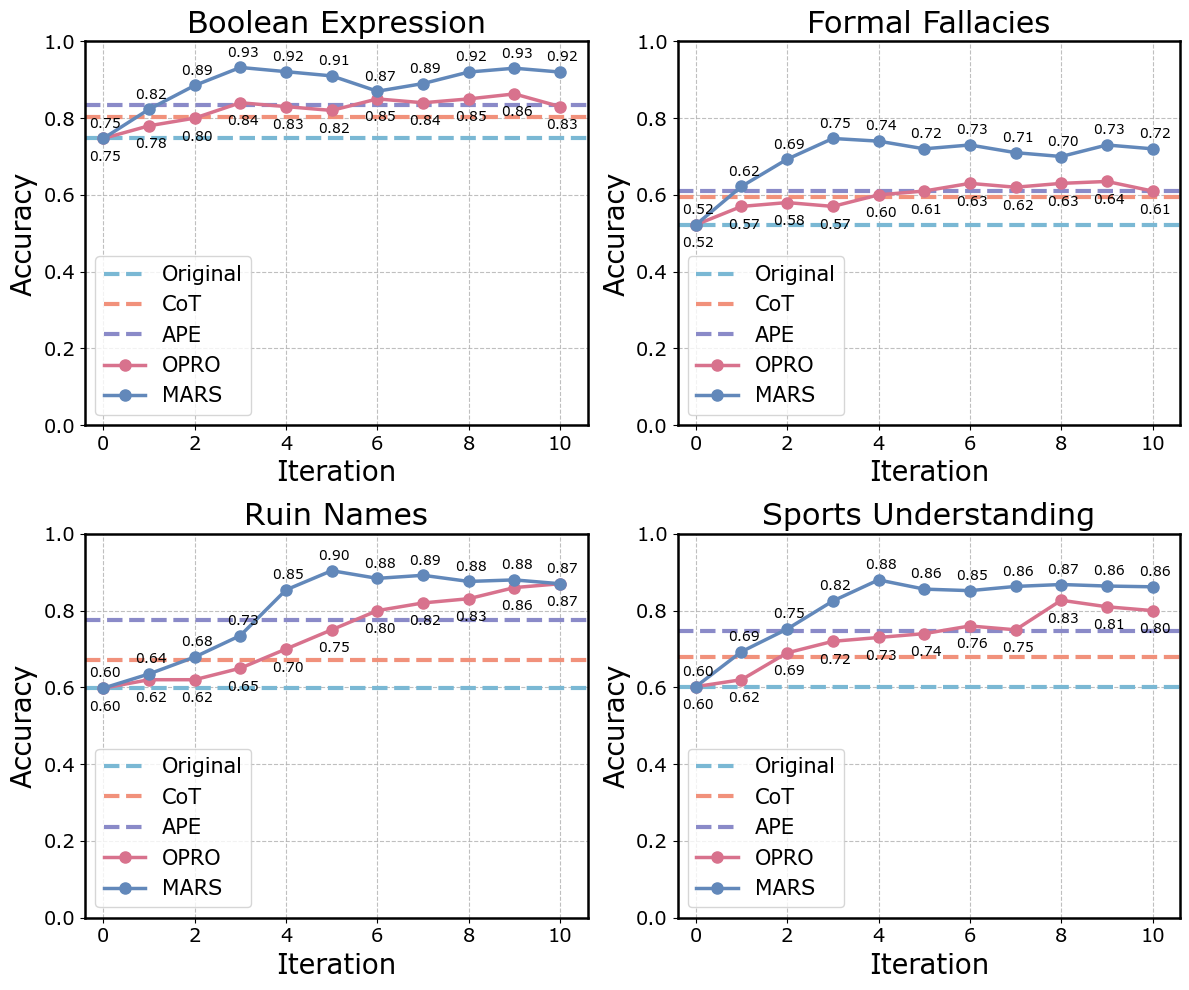}
 % \vspace{-0.6cm}
 \caption{The convergence curves across different tasks show the learning progress as the number of iterations increases. We compare the iterative convergence process of MARS with four different baseline methods across four tasks to assess MARS’s advantage in convergence speed.}
 \label{converagence}
 % \vspace{-0.4cm}
\end{figure}

\textbf{MARS achieves faster convergence in most tasks, improving both efficiency and optimization quality.} Figure~\ref{converagence} presents the convergence analysis across four BBH tasks. To better monitor the APO process, we visualize the iterative optimization trajectory within a 10-iteration observation window.

The results show that MARS exhibits an upward reward trend in the early stages. For instance, in Task `Ruin Names', it converges to the optimal solution by iteration 5. In contrast, in the OPRO task, convergence is not reached even after 10 iterations, resulting in higher resource consumption. This comparison highlights MARS’s ability to reach optimal prompts in fewer steps, reducing computational cost and enhancing efficiency.

\begin{table}[t]
\setlength{\tabcolsep}{4pt}
  \centering
  % \small % Reduce font size
  \begin{tabular}{ccc|ccc|c}
    \toprule
      \multirow{2}{*}{\textbf{Base}} & \multicolumn{2}{c|}{\textbf{Deepseek}} & \multicolumn{3}{c|}{\textbf{GPT}} 
      & \textbf{Avg.} \\
       & -V2.5 & -R1 & -3.5 & -4   & -4o
       & \\ 
    \midrule
      Origin & 56.96 & 61.48 & 44.79 & 49.70 & 55.84 & 53.75 \\[2pt]
      CoT(ZS) & 62.72 & 73.82 & 63.45 & 66.94 & 70.38 & 67.46 \\
    \midrule
      MARS & \textbf{79.59} & \textbf{83.05} & \textbf{69.30} & \textbf{73.21} & \textbf{80.86} & \textbf{77.20} \\
    \bottomrule
  \end{tabular}
  \caption{Performance comparison on BBH tasks under different \textit{Target} model settings.}
   \label{generalization}
\end{table}

\subsection{Other \textit{Target} LLMs}

% \textbf{MARS demonstrates strong cross-model generalization, maintaining high performance across diverse LLM backbones.} We further evaluate MARS on additional \textit{Target} LLMs using the optimized prompts obtained in earlier experiments. These evaluations include Deepseek-R1, GPT-3.5, GPT-4, and GPT-4o, aiming to test the method’s robustness across different model families. As shown in Table~\ref{generalization}, the prompts optimized on the Deepseek-V2.5 base model generalize effectively, preserving strong performance even on more powerful or structurally different LLMs. MARS consistently achieves notable gains across models, validating its model-agnostic design and effectiveness. This highlights MARS’s broad applicability and confirms that its optimization is not limited to a specific base model.
\textbf{MARS demonstrates strong cross-model generalization, maintaining high performance across diverse LLM backbones.} We further evaluate MARS on additional \textit{Target} LLMs—Deepseek-R1, GPT-3.5, GPT-4, and GPT-4o—using the optimized prompts from previous experiments to assess robustness across model families. As shown in Table~\ref{generalization}, prompts optimized on the Deepseek-V2.5 base model generalize well, preserving strong performance even on larger or structurally different LLMs. MARS consistently achieves notable gains across models, validating its model-agnostic design and broad applicability.

\section{Related Works}

The related work is structured into two main aspects: first, an introduction to prompt optimization; and second, an exploration of multi-agent techniques.

\paragraph{Prompt Optimization.}
Early work primarily focused on two aspects: discrete optimization of hard prompts~\citep{shin2020autoprompt,wen2024hard,chen2023instructzero,zhang2022tempera} and continuous vector optimization of soft prompts~\citep{lester2021power,li2021prefix,liu2024gpt}. However, these methods are highly task-dependent and exhibit locality. With the advent of LLMs, traditional methods have become outdated. APE~\citep{zhoularge} pioneered the use of generative methods to optimize instructions. Since APE, there have been two major approaches. The first approach~\citep{zhoularge, xureprompting,pryzant2023automatic,wang2023promptagent} is the \emph{generate-search} model, where multiple candidate sequences are generated, and methods like Monte Carlo search are used to optimize the prompt. The second approach~\citep{Yang0LLLZC24, ye2023prompt, zhang2025mars} is the \emph{meta prompts} method, where sophisticated \emph{meta prompts} are designed to optimize the prompt. In contrast to these two approaches, MARS employs a planned optimization path, iteratively generating high-quality prompts. This approach alleviates the inefficient search in prompt spaces issues in the first approach and addresses the challenges of limited flexibility of fixed templates in the second approach.

\paragraph{Multi-Agent.} Based on LLMs, a combination of AI agents capable of performing specific functions forms a multi-agent system ~\citep{richards2023auto,yang2023auto,wu2023autogen, zhang2025gkg}. Given a statement of a specific task, AI agents can attempt to break complex problem statements into subtasks and use tools, including data retrieval from the internet, to solve them step-by-step through automatic iterations. Some studies~\citep{poldrack2023ai,wang2024survey,xi2025rise,ni2021deep,lin2025foundation} use multi-agent systems to address issues such as problem identification, code development and debugging, plotting results and analysis, and providing interactive feedback with the human user.~\citet{ni2024mechagents}~\citep{zhang2025maps} demonstrates the potential of organizing an AI multi-agent collaborative team to automatically solve mechanical problems, showcasing an enhanced ability to understand, formulate, and validate engineering problem solutions through self-correction and mutual correction. Inspired by their work, we leverage multi-agent technology to autonomously plan the APO optimization path and design a \textit{Teacher}-\textit{Critic}-\textit{Student} collaborative approach for iterative optimization.

\section{Conclusion}

We propose \textbf{MARS}, a novel multi-agent framework for adaptive APO that integrates Socratic guidance within a POMDP formulation. It includes: (1) a \textit{Planner} that generates task-specific optimization trajectories, and (2) a \textit{Teacher-Critic-Student} dialogue enabling interpretable prompt refinement. This simulates pseudo-gradient paths in discrete prompt space, narrowing the search scope. Modeled as a POMDP: the \textit{Student}'s latent state is the hidden state, \textit{Teacher-Critic} interactions define actions, and prompt outputs serve as observations. A \textit{Target} agent guides iteration via performance rewards. Experiments show MARS consistently outperforms baselines while maintaining transparent optimization trajectories.

\section{Acknowledgments}

This work was supported by the National Natural Science Foundation of China (No. 62137002, 62277042, 62293553, 62450005, 62437002, 62477036, 62477037, 62176209, 62192781, 62306229), the ``LENOVO-XJTU" Intelligent Industry Joint Laboratory Project, the Shaanxi Provincial Social Science Foundation Project (No. 2024P041), the Natural Science Basic Research Program of Shaanxi (No. 2023-JC-YB-593), and the Youth Innovation Team of Shaanxi Universities ``Multi-modal Data Mining and Fusion".

\bibliography{aaai2026}

\begin{thebibliography}{39}
\providecommand{\natexlab}[1]{#1}

\bibitem[{Achiam et~al.(2023)Achiam, Adler, Agarwal, Ahmad, Akkaya, Aleman, Almeida, Altenschmidt, Altman, Anadkat et~al.}]{achiam2023gpt}
Achiam, J.; Adler, S.; Agarwal, S.; Ahmad, L.; Akkaya, I.; Aleman, F.~L.; Almeida, D.; Altenschmidt, J.; Altman, S.; Anadkat, S.; et~al. 2023.
\newblock Gpt-4 technical report.
\newblock \emph{arXiv preprint arXiv:2303.08774}.

\bibitem[{Chen et~al.(2023)Chen, Chen, Goldstein, Huang, and Zhou}]{chen2023instructzero}
Chen, L.; Chen, J.; Goldstein, T.; Huang, H.; and Zhou, T. 2023.
\newblock Instructzero: Efficient instruction optimization for black-box large language models.
\newblock \emph{arXiv preprint arXiv:2306.03082}.

\bibitem[{Guo et~al.(2025)Guo, Yang, Zhang, Song, Zhang, Xu, Zhu, Ma, Wang, Bi et~al.}]{guo2025deepseek}
Guo, D.; Yang, D.; Zhang, H.; Song, J.; Zhang, R.; Xu, R.; Zhu, Q.; Ma, S.; Wang, P.; Bi, X.; et~al. 2025.
\newblock Deepseek-r1: Incentivizing reasoning capability in llms via reinforcement learning.
\newblock \emph{arXiv preprint arXiv:2501.12948}.

\bibitem[{Huang et~al.(2024)Huang, Bai, Zhu, Zhang, Zhang, Su, Liu, Lv, Zhang, Fu et~al.}]{huang2024c}
Huang, Y.; Bai, Y.; Zhu, Z.; Zhang, J.; Zhang, J.; Su, T.; Liu, J.; Lv, C.; Zhang, Y.; Fu, Y.; et~al. 2024.
\newblock C-eval: A multi-level multi-discipline chinese evaluation suite for foundation models.
\newblock \emph{Advances in Neural Information Processing Systems}, 36.

\bibitem[{Lester, Al-Rfou, and Constant(2021)}]{lester2021power}
Lester, B.; Al-Rfou, R.; and Constant, N. 2021.
\newblock The power of scale for parameter-efficient prompt tuning.
\newblock \emph{arXiv preprint arXiv:2104.08691}.

\bibitem[{Li and Liang(2021)}]{li2021prefix}
Li, X.~L.; and Liang, P. 2021.
\newblock Prefix-tuning: Optimizing continuous prompts for generation.
\newblock \emph{arXiv preprint arXiv:2101.00190}.

\bibitem[{Lin et~al.(2025{\natexlab{a}})Lin, Zhao, He, Peng, Xu, Huang, Ma, and Feng}]{lin2025self}
Lin, Q.; Zhao, T.; He, K.; Peng, Z.; Xu, F.; Huang, L.; Ma, J.; and Feng, M. 2025{\natexlab{a}}.
\newblock Self-supervised Quantized Representation for Seamlessly Integrating Knowledge Graphs with Large Language Models.
\newblock \emph{arXiv preprint arXiv:2501.18119}.

\bibitem[{Lin et~al.(2025{\natexlab{b}})Lin, Zhu, Pu, Huang, Luo, Ma, Peng, Zhao, Xu, Zhang et~al.}]{lin2025foundation}
Lin, Q.; Zhu, Y.; Pu, B.; Huang, L.; Luo, H.; Ma, J.; Peng, Z.; Zhao, T.; Xu, F.; Zhang, J.; et~al. 2025{\natexlab{b}}.
\newblock A Foundation Model for Chest X-ray Interpretation with Grounded Reasoning via Online Reinforcement Learning.
\newblock \emph{arXiv preprint arXiv:2509.03906}.

\bibitem[{Liu et~al.(2024)Liu, Zheng, Du, Ding, Qian, Yang, and Tang}]{liu2024gpt}
Liu, X.; Zheng, Y.; Du, Z.; Ding, M.; Qian, Y.; Yang, Z.; and Tang, J. 2024.
\newblock GPT understands, too.
\newblock \emph{AI Open}, 5: 208--215.

\bibitem[{Ni and Buehler(2024)}]{ni2024mechagents}
Ni, B.; and Buehler, M.~J. 2024.
\newblock MechAgents: Large language model multi-agent collaborations can solve mechanics problems, generate new data, and integrate knowledge.
\newblock \emph{Extreme Mechanics Letters}, 67: 102131.

\bibitem[{Ni and Gao(2021)}]{ni2021deep}
Ni, B.; and Gao, H. 2021.
\newblock A deep learning approach to the inverse problem of modulus identification in elasticity.
\newblock \emph{Mrs Bulletin}, 46: 19--25.

\bibitem[{Poldrack, Lu, and Begu{\v{s}}(2023)}]{poldrack2023ai}
Poldrack, R.~A.; Lu, T.; and Begu{\v{s}}, G. 2023.
\newblock AI-assisted coding: Experiments with GPT-4.
\newblock \emph{arXiv preprint arXiv:2304.13187}.

\bibitem[{Pryzant et~al.(2023)Pryzant, Iter, Li, Lee, Zhu, and Zeng}]{pryzant2023automatic}
Pryzant, R.; Iter, D.; Li, J.; Lee, Y.~T.; Zhu, C.; and Zeng, M. 2023.
\newblock Automatic prompt optimization with "gradient descent" and beam search.
\newblock In \emph{Proceedings of the 2023 Conference on Empirical Methods in Natural Language Processing, (EMNLP)}, 7957--7968.

\bibitem[{Richards(2023)}]{richards2023auto}
Richards, T.~B. 2023.
\newblock Auto-GPT: An experimental open-source attempt to make GPT-4 fully autonomous.

\bibitem[{Shen et~al.(2025)Shen, Mao, Wang, Zhang, and Cambria}]{shentie2025}
Shen, T.; Mao, R.; Wang, J.; Zhang, X.; and Cambria, E. 2025.
\newblock Flow-guided Direct Preference Optimization for Knowledge Graph Reasoning with Trees.
\newblock In \emph{Proceedings of the 48th International ACM SIGIR Conference on Research and Development in Information Retrieval}, SIGIR '25, 1165–1175. New York, NY, USA: Association for Computing Machinery.
\newblock ISBN 9798400715921.

\bibitem[{Shin et~al.(2020)Shin, Razeghi, Logan~IV, Wallace, and Singh}]{shin2020autoprompt}
Shin, T.; Razeghi, Y.; Logan~IV, R.~L.; Wallace, E.; and Singh, S. 2020.
\newblock Autoprompt: Eliciting knowledge from language models with automatically generated prompts.
\newblock \emph{arXiv preprint arXiv:2010.15980}.

\bibitem[{Suzgun et~al.(2022)Suzgun, Scales, Sch{\"a}rli, Gehrmann, Tay, Chung, Chowdhery, Le, Chi, Zhou et~al.}]{suzgun2022challenging}
Suzgun, M.; Scales, N.; Sch{\"a}rli, N.; Gehrmann, S.; Tay, Y.; Chung, H.~W.; Chowdhery, A.; Le, Q.~V.; Chi, E.~H.; Zhou, D.; et~al. 2022.
\newblock Challenging big-bench tasks and whether chain-of-thought can solve them.
\newblock \emph{arXiv preprint arXiv:2210.09261}.

\bibitem[{Wang et~al.(2024{\natexlab{a}})Wang, Ma, Feng, Zhang, Yang, Zhang, Chen, Tang, Chen, Lin et~al.}]{wang2024survey}
Wang, L.; Ma, C.; Feng, X.; Zhang, Z.; Yang, H.; Zhang, J.; Chen, Z.; Tang, J.; Chen, X.; Lin, Y.; et~al. 2024{\natexlab{a}}.
\newblock A survey on large language model based autonomous agents.
\newblock \emph{Frontiers of Computer Science}, 18(6): 186345.

\bibitem[{Wang et~al.(2023)Wang, Li, Wang, Bai, Luo, Zhang, Jojic, Xing, and Hu}]{wang2023promptagent}
Wang, X.; Li, C.; Wang, Z.; Bai, F.; Luo, H.; Zhang, J.; Jojic, N.; Xing, E.~P.; and Hu, Z. 2023.
\newblock Promptagent: Strategic planning with language models enables expert-level prompt optimization.
\newblock \emph{arXiv preprint arXiv:2310.16427}.

\bibitem[{Wang et~al.(2024{\natexlab{b}})Wang, Ma, Zhang, Ni, Chandra, Guo, Ren, Arulraj, He, Jiang et~al.}]{wang2024mmlu}
Wang, Y.; Ma, X.; Zhang, G.; Ni, Y.; Chandra, A.; Guo, S.; Ren, W.; Arulraj, A.; He, X.; Jiang, Z.; et~al. 2024{\natexlab{b}}.
\newblock Mmlu-pro: A more robust and challenging multi-task language understanding benchmark.
\newblock \emph{arXiv preprint arXiv:2406.01574}.

\bibitem[{Wen et~al.(2024)Wen, Jain, Kirchenbauer, Goldblum, Geiping, and Goldstein}]{wen2024hard}
Wen, Y.; Jain, N.; Kirchenbauer, J.; Goldblum, M.; Geiping, J.; and Goldstein, T. 2024.
\newblock Hard prompts made easy: Gradient-based discrete optimization for prompt tuning and discovery.
\newblock \emph{Advances in Neural Information Processing Systems}, 36.

\bibitem[{Wu et~al.(2023)Wu, Bansal, Zhang, Wu, Zhang, Zhu, Li, Jiang, Zhang, and Wang}]{wu2023autogen}
Wu, Q.; Bansal, G.; Zhang, J.; Wu, Y.; Zhang, S.; Zhu, E.; Li, B.; Jiang, L.; Zhang, X.; and Wang, C. 2023.
\newblock Autogen: Enabling next-gen llm applications via multi-agent conversation framework.
\newblock \emph{arXiv preprint arXiv:2308.08155}.

\bibitem[{Xi et~al.(2025)Xi, Chen, Guo, He, Ding, Hong, Zhang, Wang, Jin, Zhou et~al.}]{xi2025rise}
Xi, Z.; Chen, W.; Guo, X.; He, W.; Ding, Y.; Hong, B.; Zhang, M.; Wang, J.; Jin, S.; Zhou, E.; et~al. 2025.
\newblock The rise and potential of large language model based agents: A survey.
\newblock \emph{Science China Information Sciences}, 68(2): 121101.

\bibitem[{Xu et~al.(2024)Xu, Wu, Sun, Ren, Yuan, Yuan, Lin, Qiao, and Liu}]{xu2023symbol}
Xu, F.; Wu, Z.; Sun, Q.; Ren, S.; Yuan, F.; Yuan, S.; Lin, Q.; Qiao, Y.; and Liu, J. 2024.
\newblock Symbol-LLM: Towards foundational symbol-centric interface for large language models.
\newblock In \emph{Proceedings of the ACL}, 13091--13116.

\bibitem[{Xu, Banburski-Fahey, and Jojic(2023)}]{xureprompting}
Xu, W.; Banburski-Fahey, A.; and Jojic, N. 2023.
\newblock Reprompting: Automated chain-of-thought prompt inference through gibbs sampling.
\newblock \emph{arXiv preprint arXiv:2305.09993}.

\bibitem[{Yan et~al.(2025)Yan, Xu, Xu, Li, Zhang, Luo, Wu, Tuan, Zhao, Lin et~al.}]{yan2025mur}
Yan, H.; Xu, F.; Xu, R.; Li, Y.; Zhang, J.; Luo, H.; Wu, X.; Tuan, L.~A.; Zhao, H.; Lin, Q.; et~al. 2025.
\newblock Mur: Momentum uncertainty guided reasoning for large language models.
\newblock \emph{arXiv preprint arXiv:2507.14958}.

\bibitem[{Yang et~al.(2024{\natexlab{a}})Yang, Wang, Lu, Liu, Le, Zhou, and Chen}]{Yang0LLLZC24}
Yang, C.; Wang, X.; Lu, Y.; Liu, H.; Le, Q.~V.; Zhou, D.; and Chen, X. 2024{\natexlab{a}}.
\newblock Large Language Models as Optimizers.
\newblock In \emph{The Twelfth International Conference on Learning Representations, {ICLR} 2024, Vienna, Austria, May 7-11, 2024}. OpenReview.net.

\bibitem[{Yang, Yue, and He(2023)}]{yang2023auto}
Yang, H.; Yue, S.; and He, Y. 2023.
\newblock Auto-gpt for online decision making: Benchmarks and additional opinions.
\newblock \emph{arXiv preprint arXiv:2306.02224}.

\bibitem[{Yang et~al.(2024{\natexlab{b}})Yang, Jin, Tang, Han, Feng, Jiang, Zhong, Yin, and Hu}]{yang2024harnessing}
Yang, J.; Jin, H.; Tang, R.; Han, X.; Feng, Q.; Jiang, H.; Zhong, S.; Yin, B.; and Hu, X. 2024{\natexlab{b}}.
\newblock Harnessing the power of llms in practice: A survey on chatgpt and beyond.
\newblock \emph{ACM Transactions on Knowledge Discovery from Data}, 18(6): 1--32.

\bibitem[{Ye et~al.(2023)Ye, Axmed, Pryzant, and Khani}]{ye2023prompt}
Ye, Q.; Axmed, M.; Pryzant, R.; and Khani, F. 2023.
\newblock Prompt engineering a prompt engineer.
\newblock \emph{arXiv preprint arXiv:2311.05661}.

\bibitem[{Yuan et~al.(2025)Yuan, Cai, Shen, Li, Huang, Deng, and Wang}]{ijcai2025p772}
Yuan, L.; Cai, Y.; Shen, X.; Li, Q.; Huang, Q.; Deng, Z.; and Wang, T. 2025.
\newblock Collaborative Multi-LoRA Experts with Achievement-based Multi-Tasks Loss for Unified Multimodal Information Extraction.
\newblock In Kwok, J., ed., \emph{Proceedings of the Thirty-Fourth International Joint Conference on Artificial Intelligence, {IJCAI-25}}, 6940--6948. International Joint Conferences on Artificial Intelligence Organization.
\newblock Main Track.

\bibitem[{Zhang et~al.(2024{\natexlab{a}})Zhang, Da, Lee, Robinson, Wu, Song, Zhao, Raja, Slack, Lyu et~al.}]{zhang2024careful}
Zhang, H.; Da, J.; Lee, D.; Robinson, V.; Wu, C.; Song, W.; Zhao, T.; Raja, P.; Slack, D.; Lyu, Q.; et~al. 2024{\natexlab{a}}.
\newblock A careful examination of large language model performance on grade school arithmetic.
\newblock \emph{arXiv preprint arXiv:2405.00332}.

\bibitem[{Zhang et~al.(2025{\natexlab{a}})Zhang, Wang, Wang, Zhang, Xu, Lin, Mao, Cambria, and Liu}]{zhang2025maps}
Zhang, J.; Wang, Z.; Wang, Z.; Zhang, X.; Xu, F.; Lin, Q.; Mao, R.; Cambria, E.; and Liu, J. 2025{\natexlab{a}}.
\newblock MAPS: A Multi-Agent Framework Based on Big Seven Personality and Socratic Guidance for Multimodal Scientific Problem Solving.
\newblock \emph{arXiv preprint arXiv:2503.16905}.

\bibitem[{Zhang et~al.(2025{\natexlab{b}})Zhang, Wang, Zhu, Liu, Lin, and Cambria}]{zhang2025mars}
Zhang, J.; Wang, Z.; Zhu, H.; Liu, J.; Lin, Q.; and Cambria, E. 2025{\natexlab{b}}.
\newblock MARS: A Multi-Agent Framework Incorporating Socratic Guidance for Automated Prompt Optimization.
\newblock \emph{arXiv preprint arXiv:2503.16874}.

\bibitem[{Zhang et~al.(2025{\natexlab{c}})Zhang, Wei, Qi, Liu, Lin et~al.}]{zhang2025gkg}
Zhang, J.; Wei, B.; Qi, S.; Liu, J.; Lin, Q.; et~al. 2025{\natexlab{c}}.
\newblock GKG-LLM: A Unified Framework for Generalized Knowledge Graph Construction.
\newblock \emph{arXiv preprint arXiv:2503.11227}.

\bibitem[{Zhang et~al.(2024{\natexlab{b}})Zhang, Yang, Zhu, Lin, Xu, and Liu}]{zhang2024semantic}
Zhang, J.; Yang, C.; Zhu, H.; Lin, Q.; Xu, F.; and Liu, J. 2024{\natexlab{b}}.
\newblock A Semantic Mention Graph Augmented Model for Document-Level Event Argument Extraction.
\newblock \emph{arXiv preprint arXiv:2403.09721}.

\bibitem[{Zhang et~al.(2022)Zhang, Wang, Zhou, Schuurmans, and Gonzalez}]{zhang2022tempera}
Zhang, T.; Wang, X.; Zhou, D.; Schuurmans, D.; and Gonzalez, J.~E. 2022.
\newblock Tempera: Test-time prompting via reinforcement learning.
\newblock \emph{arXiv preprint arXiv:2211.11890}.

\bibitem[{Zhong et~al.(2023)Zhong, Cui, Guo, Liang, Lu, Wang, Saied, Chen, and Duan}]{zhong2023agieval}
Zhong, W.; Cui, R.; Guo, Y.; Liang, Y.; Lu, S.; Wang, Y.; Saied, A.; Chen, W.; and Duan, N. 2023.
\newblock Agieval: A human-centric benchmark for evaluating foundation models.
\newblock \emph{arXiv preprint arXiv:2304.06364}.

\bibitem[{Zhou et~al.(2022)Zhou, Muresanu, Han, Paster, Pitis, Chan, and Ba}]{zhoularge}
Zhou, Y.; Muresanu, A.~I.; Han, Z.; Paster, K.; Pitis, S.; Chan, H.; and Ba, J. 2022.
\newblock Large language models are human-level prompt engineers.
\newblock \emph{arXiv preprint arXiv:2211.01910}.

\end{thebibliography}

\clearpage

\section{Proof of Proposition}

\subsection{Proof of Proposition 1 (Socratic Policy Improvement Bound)}

Let \(\Pi = \{\pi_{\text{t}}, \pi_{\text{c}}, \pi_{\text{s}}\}\) denote the joint policy composed of the Teacher, Critic, and Student agents. Let \(p_i\) denote the prompt state at refinement step \(i\), and define the associated latent state as \(s_i\). At each step, the action \(a_i = (q_i, c_i)\) induces a transition from \(s_{i-1}\) to \(s_i\), and yields a prompt \(p_i = \pi_{\text{s}}(a_i, p_{i-1})\).

We assume the existence of an underlying task reward function:
\begin{equation}
\mathcal{R}(p) = \mathbb{E}_{(x, y) \sim D} \left[ f\left( \pi_{\text{tar}}(x; p), y \right) \right],
\end{equation}
which is Lipschitz-continuous with constant \(\lambda > 0\). Our goal is to lower-bound the cumulative reward improvement from \(p_0\) to \(p_n\) based on Socratic signal quality.

\paragraph{Step 1: Define Local Advantage.}
Let the local reward gain be defined as:
\begin{equation}
A_i := \mathcal{R}(p_i) - \mathcal{R}(p_{i-1}).
\end{equation}
We assume the expected advantage at each step satisfies:
\begin{equation}
\mathbb{E}[A_i] \ge \bar{A}_i > 0,
\end{equation}
i.e., the expected contribution of the composite action \(a_i\) is beneficial.

\paragraph{Step 2: Control for Socratic Variance.}
We define the conditional variance of \(A_i\) given the state \(s_{i-1}\) as:
\begin{equation}
\text{Var}[A_i \mid s_{i-1}] \le \sigma^2.
\end{equation}
This variance captures uncertainty due to imperfect questions or noisy feedback in the Socratic interaction.

\paragraph{Step 3: Apply Jensen-Bernstein Inequality.}
Since \(\mathcal{R}(p)\) is Lipschitz, and prompt updates occur in a discrete space, we use a Bernstein-style bound:
\begin{equation}
\mathbb{E}[\mathcal{R}(p_i)] \ge \mathcal{R}(p_{i-1}) + \bar{A}_i - \frac{\sigma^2}{2\lambda}.
\end{equation}

\paragraph{Step 4: Accumulate Across Steps.}
Summing over all \(n\) refinement steps, we obtain:
\begin{equation}
\begin{aligned}
\mathbb{E}[\mathcal{R}(p_n)] - \mathcal{R}(p_0)
&= \sum_{i=1}^n \mathbb{E}[A_i] \\
&\ge \sum_{i=1}^n \left( \bar{A}_i - \frac{\sigma^2}{2\lambda} \right).
\end{aligned}
\end{equation}

This completes the proof of the lower bound. It shows that as long as the Socratic actions are informative (\(\bar{A}_i > 0\)) and the variance \(\sigma^2\) is controlled, the cumulative reward is guaranteed to improve linearly with the number of refinement steps.

\subsection{Proof of Proposition 2 (Monotonic Reward Stability)}

Let \(\{p_i\}_{i=0}^n\) denote the sequence of prompts generated by the MARS refinement process, where each prompt is updated via:
\begin{equation}
p_i = \pi_{\text{s}}(a_i, p_{i-1}), \quad \text{with} \quad a_i = (q_i, c_i),
\end{equation}
i.e., a Socratic action \(a_i\) is used to refine the prompt at step \(i\).

Assume the reward function \(\mathcal{R}: \mathcal{P} \rightarrow \mathbb{R}\) is \(\lambda\)-Lipschitz continuous with respect to the prompt representation, i.e., for all \(i = 1, \dots, n\),
\begin{equation}
|\mathcal{R}(p_i) - \mathcal{R}(p_{i-1})| \le \lambda \cdot \|p_i - p_{i-1}\|,
\end{equation}
where \(\|\cdot\|\) denotes a norm over the prompt space (e.g., token-level edit distance or embedding-based distance).

Suppose further that the update step size is bounded as:
\begin{equation}
\|p_i - p_{i-1}\| \le \varepsilon.
\end{equation}
Then, combining (2) and (3), the reward change per step is bounded by:
\begin{equation}
|\mathcal{R}(p_i) - \mathcal{R}(p_{i-1})| \le \lambda \varepsilon.
\end{equation}

This implies that the reward improvement (or degradation) at each refinement step is at most linear in the prompt change size.

Now suppose the process satisfies a minimum improvement requirement:
\begin{equation}
\mathcal{R}(p_i) \ge \mathcal{R}(p_{i-1}) + \delta,
\end{equation}
for some \(\delta > 0\). Then combining (4) and (5), we obtain:
\begin{equation}
\lambda \varepsilon \ge \delta \quad \Rightarrow \quad \varepsilon \ge \delta / \lambda.
\end{equation}

Thus, for a reward gain of at least \(\delta\), the update must induce a prompt change of at least \(\delta / \lambda\). Conversely, if
\begin{equation}
\|p_i - p_{i-1}\| \ll \delta / \lambda,
\end{equation}
then the step is too small to yield meaningful improvement, and further refinement is unlikely to be effective.

This result provides a formal justification for the early stopping rule: once consecutive prompt updates fall below a semantic change threshold, reward improvement will necessarily be bounded, indicating convergence.

\section{Tasks and Datasets}
\label{appendixA}
To comprehensively evaluate the expert-level prompt optimization capabilities of our framework, we curate 17 tasks from two broad categories: General Tasks and Domain-Specific Tasks.

\paragraph{General Task Evaluation.} We select six tasks from the BBH~\citep{suzgun2022challenging} and MMLU~\citep{wang2024mmlu} datasets, respectively. BBH tasks consist of six challenging reasoning tasks that assess logical inference and problem-solving skills, including boolean expressions, disambiguation QA, formal fallacies, geometric shapes, ruin names, and sports understanding. MMLU tasks cover six subject-specific tasks designed to evaluate general knowledge across diverse fields, including college biology, college medicine, electrical engineering, high school world history, human aging, and marketing.

\paragraph{Domain-Specific Task Evaluation.} We include three benchmarks: C-Eval~\citep{huang2024c}, GSM8K~\citep{zhang2024careful}, and LSAT-AR~\citep{zhong2023agieval}. C-Eval is a Chinese evaluation benchmark that covers domain-specific topics such as art studies, clinical medicine, and Urban and Rural Planner. GSM8K is a widely used mathematical reasoning dataset. LSAT-AR focuses on legal reasoning, evaluating AI performance in law-related tasks.

\paragraph{Dataset Split.}
In this study, we adopt a minimal training paradigm by selecting only a single instance from each dataset for training. Despite this extremely limited supervision, our method demonstrates strong and consistent performance across a diverse range of datasets. This suggests that our approach possesses exceptional few-shot ability, enabling effective adaptation to various tasks with minimal prior knowledge. The detailed partition of the dataset is presented in Table~\ref{datasetsDetails}.

One of the key highlights of this study is that the training data consists of only a single sample—MARS utilizes just one data point from the current task for the entire optimization process. This minimal setup is enabled by the \textit{Planner} agent’s strong capacity to identify the task definition and interpret the provided example, allowing it to infer the underlying task structure and semantics effectively. Since the primary function of the \textit{Planner} is to generate an optimization trajectory based on the prompt-task alignment, a single representative instance is sufficient to guide downstream agents. We further analyze the impact of different sample sizes on performance in Appendix~E.

\begin{table}[t]
\setlength{\tabcolsep}{8pt}
  \centering
  \small % Reduce font size
  \begin{tabular}{lccc}
    \toprule
      Tasks & ABBR. & Train & Test\\
    \midrule
      \textbf{Bigbench} & ~ & ~ & ~\\[2pt]
      \hspace{0.5cm}Boolean Expressions & B.E. & 1 & 249 \\[2pt] 
      \hspace{0.5cm}Disambiguation QA & D.QA & 1 & 249 \\ [2pt]
      \hspace{0.5cm}Formal Fallacies & F.F. & 1 & 249 \\ [2pt]
      \hspace{0.5cm}Geometric Shapes & G.S. & 1 & 249 \\ [2pt]
      \hspace{0.5cm}Ruin Names &R.N. & 1 & 249 \\ [2pt]
      \hspace{0.5cm}Sports Understanding & S.U. & 1 & 249 \\[2pt] 
      \textbf{MMLU} & ~ & ~ & ~\\[2pt]
      \hspace{0.5cm}College Biology & C.B. & 1 & 143 \\[2pt]
      \hspace{0.5cm}College Medicine & C.M. & 1 & 172 \\[2pt]
      \hspace{0.5cm}Electrical Engineering & E.E. & 1 & 144 \\[2pt]
      \hspace{0.5cm}HighSchool World History & W.H. & 1 & 236 \\[2pt]
      \hspace{0.5cm}Human Aging & H.A. & 1 & 222 \\[2pt]
      \hspace{0.5cm}Marketing & M.T. & 1 & 233 \\
    \midrule
      \textbf{C-EVAL} & ~ & ~ & ~ \\[2pt]

      \hspace{0.5cm}Art Studies & A.S. & 1 & 32 \\[2pt]
      \hspace{0.5cm}Urban And Rural Planner & U.R.P. & 1 & 45 \\[2pt]
      \hspace{0.5cm}Clinical Medicine & CL.M. & 1 & 21 \\[2pt]
      \textbf{GSM8K} & GSM. & 1 & 1318 \\[2pt]
      \textbf{LSAT-AR} & L.A. & 1 & 229 \\
    \bottomrule
  \end{tabular}
  \caption{Data split of general tasks and domain-specific tasks. One instance for training and others for testing. The `ABBR.' column represents the abbreviations for all the tasks.}
  \label{datasetsDetails}
\end{table}
% \FloatBarrier % 确保表格不会漂移到其他章节

\section{Experiment Settings and Baselines} \label{appendixAA}

We select a powerful LLM, deepseek-V2.5-1210~\citep{guo2025deepseek}, as our primary agent for the APO tasks. Not only does deepseek-V2.5-1210 exhibit strong reasoning and generation capabilities in a variety of natural language processing tasks, but it also efficiently explores multiple angles when facing complex prompt optimization requirements, making it well-suited for adapting to different tasks and datasets in the APO process. We adopt accuracy as our primary evaluation metric to comprehensively assess the performance of different methods across various task scenarios. 
Table 1 presents our experimental results on 12 general task datasets, illustrating the performance of APO in diverse scenarios, while Table 2 summarizes its performance on five domain-specific datasets, underscoring the model's versatility and stability across different fields. To further validate the generality and robustness of our method, we additionally employed another high-performance LLM, GPT-4o~\citep{achiam2023gpt}, for extended comparative experiments, with the corresponding findings reported in Appendix D.

We compare MARS with three categories of baselines: original prompts, CoT prompts, and some of the latest APO methods. Specifically, (1) original prompts refer to the prompts used in the datasets, where each dataset often provides some initial guidance for the tasks. (2) To build the CoT (Zero-Shot) baseline, we add the prompt \textit{Let's think step by step} at the beginning of each task; based on this, we further include a specific example to create the CoT (Few-Shot) baseline. (3) Finally, we compare MARS with some strong baseline methods from recent years, including Automatic Prompt Engineer (APE)~\citep{zhoularge}, Prompt Optimization with Textual Gradients (ProTeGi)~\citep{pryzant2023automatic}, Optimization by PROmpting (OPRO)~\citep{Yang0LLLZC24}, and Prompt Engineer 2 (PE2)~\citep{ye2023prompt}. APE and ProTeGi generate multiple prompts and perform search optimization to find the optimal prompt, while OPRO and PE2 optimize prompts by designing a sophisticated \emph{meta prompts}.

\section{Generalization Across Different Base and Target Models}
\label{appendixB}

 In this section, we present the optimization performance of the method from this study on another base model, as well as the optimization results of our APO on other \textit{Target} LLMs.

\subsection{Base Model of GPT-4o}

To verify the generality and effectiveness of the proposed method in this study, we conduct further experiments by replacing the base model with GPT-4o~\citep{achiam2023gpt}.

As shown in Table~\ref{anotherBase}, in the datasets of the 17 tasks adopted by this study, MARS achieves a new SOTA performance when using the GPT-4o base model, surpassing the previous SOTA by 2.3\%. This result demonstrates that MARS not only performs excellently on the existing base models but also exhibits strong transferability, continuously improving performance across different base models. This further validates the versatility and robustness of the MARS method, highlighting its effectiveness on a variety of base models.

\subsection{Different Initial Prompts \(p_0\)}

As shown in the Table~\ref{initial}, MARS consistently achieves strong performance across different initial prompts~$p_0$, demonstrating robust optimization capabilities. Although the choice of $p_0$ can lead to variations in absolute performance, MARS maintains a relatively stable improvement margin across tasks. This indicates that MARS effectively adapts its optimization trajectory regardless of the quality of the starting prompt, highlighting its reliability and generalization ability in diverse initialization scenarios.

\begin{table}[t]
\setlength{\tabcolsep}{5pt}
  \centering
  \small % Reduce font size
  \begin{tabular}{cccccc|c}
    \toprule
      Tasks & BBH & MMLU & Chinese & GSM. & L.A. 
       & Avg. \\ 
    \midrule
      Origin & 60.92 & 83.73 & 58.26 & 72.31 & 20.96 & 59.24 \\[2pt]
      CoT(ZS) & 62.81 & 85.62 & 64.26 & 76.25 & 24.45 & 62.68 \\[2pt]
      CoT(FS) & 63.42 & 88.27 & 68.69 & 83.92 & 28.82 & 66.62 \\[2pt]
      APE   & 64.36 & 86.72 & 69.03 & 81.18 & 30.13 & 66.28 \\[2pt]
      ProTeGi & 76.43 & 86.35 & 73.52 & 82.70 & 31.88 & 70.18 \\[2pt]
      OPRO  & \underline{78.73} & 88.25 & \underline{75.79} & 84.74 & 32.75 & 72.05 \\[2pt]
      PE2 & 77.59 & \underline{91.89} & 74.67 & \underline{85.43} & \underline{35.81} & \underline{73.08} \\
    \midrule
      MARS & \textbf{81.13} & \textbf{92.82} & \textbf{78.11} & \textbf{90.97} & \textbf{40.17} & \textbf{76.58} \\
    \bottomrule
  \end{tabular}
  \caption{Performance comparison on difference tasks based on GPT-4o.}
   \label{anotherBase}
\end{table}

\begin{table}[htp]
\setlength{\tabcolsep}{9pt}
  \centering
  \small % Reduce font size
  \begin{tabular}{cp{1.3cm}p{1.2cm}p{1.2cm}|c}
    \toprule
        \textbf{Model} & \textbf{1} & \textbf{2}& \textbf{3}& \textbf{Avg}\\
    \midrule
      $p_0$  & Let's think step by step. & Let’s work this out step by step. & Let’s proceed with our tasks one by one. &    \\
    \midrule
      MARS & \textbf{79.59} & \textbf{78.53} & \textbf{75.30} & \textbf{77.81}  \\
    \bottomrule
  \end{tabular}
  \caption{Performance comparison of MARS on BBH tasks under different initial prompts \(p_0\).
}
   \label{initial}
\end{table}

\section{Sample Size Analysis} \label{appendixBB}

To analyze the rationality of one-shot training, we present a comparison of 0-shot, 1-shot, 3-shot, and baseline methods in Table~\ref{samples}.

\begin{table}[ht]
\setlength{\tabcolsep}{4pt}
  \centering
  \small % Reduce font size
  \begin{tabular}{cc|cccccc}
    \toprule
      \textbf{Tasks} & \textit{Train} & \textbf{B.E.} & \textbf{D.QA} & \textbf{F.F.} & \textbf{G.S.} & \textbf{R.N.} & \textbf{S.U.}\\
    \midrule
      % Tasks & TrainingData & Boolean Expressions & Disambiguation QA & Formal Fallacies Syllogisms Negation & Geometric Shapes & Ruin Names & Sports Understanding \\ 
      APE & 100 & 83.53 & 61.85 & 61.04 & 51.41 & 77.51 & 74.70 \\
      ProTeGi & 20 & 83.93 & 63.86 & 62.65 & 52.21 & 80.32 & 76.71 \\
      OPRO & 50 & 86.34 & 66.67 & 63.45 & 53.81 & 83.13 & 82.73 \\
      PE2 & 100 & 87.95 & 65.46 & 63.86 & 54.62 & 84.34 & 75.90 \\
    \midrule
      MARS & 0 & 90.76 & 70.28 & 73.09 & 57.83 & 88.35 & 85.94 \\ 
      MARS & 1 & 93.17 & 71.89 & 74.70 & 59.43 & 90.36 & 87.95 \\ 
      MARS & 3 & 93.57 & 72.69 & 74.30 & 60.24 & 89.96 & 88.35 \\ 
    \bottomrule
  \end{tabular} 
  \caption{Performance comparison of different sampling strategies on the evaluation metric. \textit{Train} means the training data.}
  \label{samples}
  % \vspace{-0.6cm}
\end{table}

The results indicate that the performance difference between 1-shot and 3-shot is minimal, yet the 1-shot approach is more resource-efficient while also enhancing task time efficiency. This demonstrates that in resource-constrained scenarios, 1-shot training offers a better trade-off between performance and computational cost. Other strong baseline models, such as PromptAgent and OPRO, use at least 20\% of the data for training, while our framework, using 1-shot training, achieves better performance than these models. This clearly demonstrates the effectiveness and resource efficiency of the MARS method.

\section{Prompts for Agents}
\label{appendixC}

Table~\ref{agentPrompts} summarizes the prompts used for all agents in this paper, each playing a crucial role in the overall optimization workflow.

The \textit{Planner} first constructs a structured plan based on the task requirements, defining a trajectory of sub-goals to guide the optimization process. Beyond laying out the overall flow, the \textit{Planner} provides semantic anchors that structure the interactions among downstream agents.

In the refinement phase, the \textit{Teacher} generates Socratic-style questions aligned with the current sub-goal, designed to elicit reasoning rather than direct edits. The \textit{Student} responds by proposing refined prompts, while the \textit{Critic} evaluates the quality and pedagogical alignment of the guidance, forming an interactive loop for iterative improvement.

Finally, the \textit{Target} agent validates the final prompt on downstream tasks, providing external performance feedback that closes the optimization loop. This validation ensures that the generated prompt is not only structurally coherent but also effective for the intended task.

\section{Full-process Prompt Optimization}
\label{appendixD}

\begin{figure*}[t]
	\large
	\centering

	\includegraphics[scale=0.35]{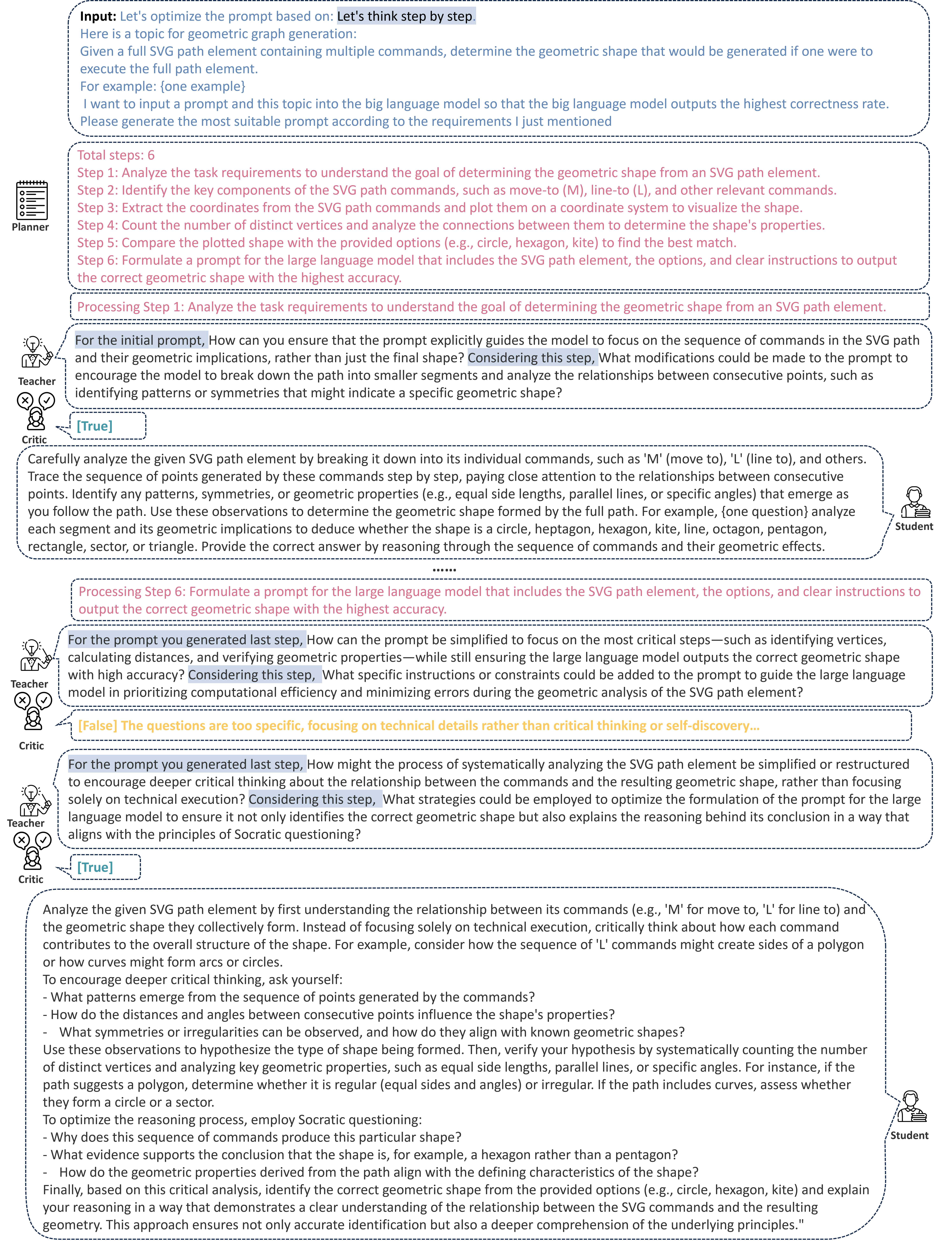}
	\caption{This figure presents a complete example of the collaborative output from all agents in a single iteration, using the Geometry Shapes task from the BBH dataset.}
	\label{fullSteps}
	% \vspace{-0.3cm}
\end{figure*}

Figure~\ref{fullSteps} presents a comprehensive example of full-process prompt optimization, using the \textit{Geometry Shapes} task from the BBH dataset. This visualization clearly illustrates the end-to-end workflow of the \textbf{MARS} framework, highlighting how the multi-agent system approximates a policy-guided trajectory over a discrete prompt space.

The process begins with the \textit{Planner} agent, which decomposes the input task into a series of interpretable sub-goals \(\{st_1, \dots, st_n\}\), forming a high-level optimization trajectory. These steps serve as a form of global guidance for the subsequent reasoning path, representing the initial policy direction in the POMDP formulation.

Then, through the iterative interaction among the \textit{Teacher}, \textit{Critic}, and \textit{Student} agents, MARS executes a sequence of transitions \((s_{i-1}, a_i, s_i)\), where each composite action \(a_i = (q_i, c_i)\) is derived from the Socratic-style question and its critique. The \textit{Student} agent updates its internal latent state \(s_i\) based on this feedback and outputs the observable prompt \(o_i = p_i\). This dialogue-driven process simulates a soft pseudo-gradient trajectory in the POMDP landscape, gradually refining the prompt through interpretable and feedback-aligned steps.

\textbf{MARS enables dual-level interpretability:} \textit{process interpretability} arises from the explicit optimization path planned by the \textit{Planner} and the Socratic dialogue structure, which makes each state transition traceable and rational. \textit{Result interpretability} is embodied in the final prompt, which integrates task-specific constraints---such as tolerance thresholds or validation rules---as shown in Figure~\ref{fullSteps}, indicating the policy-converged output under the POMDP framework.

\section{Universal Optimum Solution}
\label{appendixE}

This section introduces the final optimized prompts for all general tasks and domain-specific tasks, obtained through the MARS optimization process. After multiple iterations for each of the 17 sub-tasks, a prompt strategy that yields optimal performance was identified for each one. Tables~\ref{resultsBE} through Table~\ref{resultsCM} sequentially present the best solutions for these 17 sub-tasks along with their respective experimental results, demonstrating the adaptability and effectiveness of MARS across a broad range of tasks.

\begin{table*}[t]
% \begin{supertabular}{t}
  \renewcommand{\arraystretch}{1.2} % 增加行距
  \centering
  \small
  % \begin{tabularx{\linewidth}{>{\raggedright\arraybackslash}X} } % 让内容自动换行 + 左对齐
  \begin{tabular}{p{16cm}} 
  \toprule
  \textbf{\textit{Planner}} \\
  Split the task '{Here is a topic for geometric graph generation:
Given a full SVG path element containing multiple commands, determine the geometric shape that would be generated if one were to execute the full path element.

For example:
This SVG path element <path d=M 64.00,63.00 L 44.00,63.00 L 44.00,50.00 L 64.00,50.00 L 64.00,45.00 L 85.00,57.00 L 64.00,68.00 L 64.00,63.00""/> draws a 
Options:
(A) circle
(B) heptagon
(C) hexagon
(D) kite
(E) line
(F) octagon
(G) pentagon
(H) rectangle
(I) sector
(J) triangle

I want to input a prompt and this topic into the big language model so that the big language model outputs the highest correctness rate. 
Please generate the most suitable prompt according to the requirements I just mentioned.}' into detailed steps and details.

For example, for the clinical medicine Test, the task is planned as follows:
Total steps: 4 
Step 1: Analyze the input requirements, focusing on the type of clinical medicine question and the format of the options. 
Step 2: Design a prompt that encourages the model to consider the specific clinical characteristics of the condition described in the question and match the most appropriate option based on medical knowledge. 
Step 3: Request the model to evaluate each option in the context of clinical presentation, symptoms, and diagnostic characteristics of the condition to ensure it selects the most accurate answer. 
Step 4: Test and refine the prompt to ensure the model produces the highest correctness rate for similar clinical medicine questions.\\

  \midrule
  \textbf{\textit{Teacher}} \\
  You are a teacher who asks questions in the Socratic manner based on objectives and student responses. 
Please ask a total of two questions:
The first one is for the problem that appeared in the prompt given by the students in the last round.
The second one is an optimization solution based on the current steps of the task.

Please include only questions in your output and do not make answers for your students.\\

  \midrule
  \textbf{\textit{Student}} \\
  You are a prompt generator, please proceed to iterate over the existing prompts as required.

Note that you should only output the new prompt you generated.\\
  \midrule
  \textbf{\textit{Critic}} \\
  You are an evaluator responsible for judging the correctness of a given task. Your output must strictly follow these rules:

  1. If the task is judged as correct, output only:
    [True]
  2. If the task is judged as incorrect, output:
    [False]
    [suggestion: <reason for the incorrect judgment>]
  
    Replace `<reason for the incorrect judgment>` with a clear and concise explanation of why the task is incorrect.
  
  Do not include any additional text, comments, or explanations beyond the specified format.\\
  \midrule
  \textbf{\textit{Target}} \\
  Prompt: Systematically analyze the given SVG path element by first breaking it down into its individual commands, such as 'M' (move to), 'L' (line to), and others. For each command, map the sequence of points it generates, ensuring you accurately trace the path step by step. As you follow the path, focus on identifying key geometric properties, such as equal side lengths, parallel lines, specific angles, or symmetries, that emerge between consecutive points. Use these properties to classify the shape based on its defining characteristics. For example, given the path <path d=M 64.00,63.00 L 44.00,63.00 L 44.00,50.00 L 64.00,50.00 L 64.00,45.00 L 85.00,57.00 L 64.00,68.00 L 64.00,63.00/>, calculate the distances between points to check for equal side lengths, measure angles to identify parallelism or perpendicularity, and look for symmetries that align with known geometric shapes. Based on these observations, determine whether the shape is a circle, heptagon, hexagon, kite, line, octagon, pentagon, rectangle, sector, or triangle. Provide the correct answer by reasoning through the geometric properties derived from the path commands.

Question: {question }

Please don't output the process of doing the question, only the content of the answer.The answer should be a parenthesis containing the capital letter of the chosen answer. please do not add any other spaces or symbols.\\
  \bottomrule
  \end{tabular}
\caption{The table summarizes the prompts used for all Agents in this paper. The examples in the table are from the Geometry Shapes Task of the BBH dataset.}
\label{agentPrompts}

\end{table*}
% \end{supertabular}
% \FloatBarrier % 确保表格不会漂移到其他章节

\begin{table*}[t]
  \renewcommand{\arraystretch}{1.2} % 增加行距
  \begin{tabular}{p{16cm}} 
  \toprule
  \textbf{Boolean Expressions} \\

  Evaluate the truth value of the following Boolean expression step by step. The expression consists of Boolean constants (True, False) and basic Boolean operators (and, or, not). Carefully analyze each part of the expression, apply the correct Boolean logic, and provide the final truth value as your answer. For example, if the input is 'not ( True ) and ( True ) is', the correct output is 'False'. Ensure your reasoning is clear and accurate.\\
  \bottomrule

  \end{tabular}
  \caption{The table shows the final optimized prompt for the Boolean Expressions task of BBH using the MARS method.}
  \label{resultsBE}
\end{table*}

\begin{table*}[t]
  \renewcommand{\arraystretch}{1.2}
  \begin{tabular}{p{16cm}} 
  \toprule
  \textbf{Disambiguation QA} \\
  Analyze the following sentence to determine whether the pronoun is inherently ambiguous or if it can be linked to a specific antecedent. Follow these streamlined steps to efficiently evaluate pronoun disambiguation while maintaining accuracy, especially in complex sentence structures:

1. Identify the Pronoun and Its Grammatical Role: 

2. Identify Key Contextual Cues: 

3. List and Filter Potential Antecedents: 

4. Evaluate Plausibility: 

5. Determine Ambiguity or Specific Antecedent: 

6. Align with Provided Options: 

Evaluation Metrics for Model Output: 

1. Correctness: 

2. Clarity: 

3. Efficiency: 

4. Consistency: 

Additional Considerations: 

1. Grammatical Structure Influence: 

2. Optimizing Contextual Cue Identification: 

By simplifying the steps and focusing on key evaluation metrics, the model can process and apply the disambiguation process more efficiently while maintaining high accuracy and clarity in its outputs, even in complex sentence structures.\\
  \bottomrule
  \end{tabular}
  \caption{The table shows the final optimized prompt for the Disambiguation QA task of BBH using the MARS method.}
\end{table*}

\begin{table*}[t]
  \renewcommand{\arraystretch}{1.2}
  \begin{tabular}{p{16cm}} 
  \toprule
  \textbf{Formal Fallacies Syllogisms Negation} \\
  Analyze the following argument step by step to determine its logical validity. Carefully consider the premises provided and assess whether the conclusion necessarily follows from them. Pay special attention to the role of negations in the argument. After evaluating the logical structure, decide whether the argument is deductively valid or invalid based on the given premises. Choose the correct option from the provided choices: valid or invalid. Ensure your reasoning is thorough and aligns with formal logical principles. \\
  \bottomrule
  \end{tabular}
  \caption{The table shows the final optimized prompt for the Formal Fallacies Syllogisms Negation task of BBH using the MARS method.}
\end{table*}

\begin{table*}[t]
  \renewcommand{\arraystretch}{1.2}
  \begin{tabular}{p{16cm}} 
  \toprule
  \textbf{Geometric Shapes} \\
  Given an SVG path element and a list of geometric shape options, systematically analyze and interpret the sequence of SVG path commands to determine the number of vertices and the overall structure of the geometric shape. Follow this structured and optimized approach:

1. Dynamic Tolerance Threshold for Vertex Identification: {Detailed Explanation}

2. Optimized Vertex Counting and Connection: {Detailed Explanation}

3. Critical SVG Path Command Analysis:{Detailed Explanation}

4. Accurate Vertex Counting and Connection:{Detailed Explanation}

5. Distinguishing Between Similar Shapes:{Detailed Explanation}

6. Systematic Comparison with Provided Options:{Detailed Explanation}

7. Validation and Refinement:{Detailed Explanation}

8. Optimization for Similar Shapes:{Detailed Explanation}

Key Considerations for Dynamic Tolerance and Vertex Identification:{Detailed Explanation}

Optimized Comparison Process:{Detailed Explanation}

By integrating these considerations into the analysis, the model can achieve a higher correctness rate in identifying geometric shapes from SVG paths, even when dealing with shapes that have similar properties. \\
  \bottomrule
  \end{tabular}
  \caption{The table shows the final optimized prompt for the Geometric Shapes task of BBH using the MARS method.}
\end{table*}

\begin{table*}[t]
  \renewcommand{\arraystretch}{1.2}
  \begin{tabular}{p{16cm}} 
  \toprule
  \textbf{Ruin Names} \\
  Given an artist, band, or movie name, create a one-character edit that changes the name in a humorous and universally recognizable way. The edit must involve only a single-character change (adding, removing, or substituting one letter) and should prioritize simplicity, absurdity, and surprise to evoke humor effectively. Ensure the edit maintains a clear connection to the original name, making the humor immediately recognizable and universally understandable, while avoiding overly specific or niche references. 

Key Guidelines: 
1. Simplicity and Surprise: 
2. Cultural Universality: 
3. Absurdity and Creativity: 

Evaluation Metrics: 

Strategies for Simplicity and Surprise: 

Systematic Testing Strategies: 

Examples: 

Refinement for Evaluation Metrics: 

Focus on generating edits that are simple, surprising, and universally amusing, ensuring they strictly adhere to the one-character constraint and meet the evaluation criteria for humor, cultural relevance, and clarity. Test each edit with a diverse set of sample inputs and audiences to validate its humor consistency and cultural universality, ensuring the edit is immediately recognizable and universally understandable.\\
  \bottomrule
  \end{tabular}
  \caption{The table shows the final optimized prompt for the Ruin Names task of BBH using the MARS method.}
\end{table*}

\begin{table*}[t]
  \renewcommand{\arraystretch}{1.2}
  \begin{tabular}{p{16cm}} 
  \toprule
  \textbf{Sports Understanding} \\
  Evaluate the plausibility of the following sports-related sentence by considering the following key aspects: 

1. Player Abilities and Historical Performance: 

2. Event Context and Historical Significance: 

3. Terminology and Sport-Specific Knowledge: 

4. Rarity vs. Impossibility: 

Guidelines: 
- If the action is rare but historically documented or consistent with the player's abilities, consider it plausible. 
- For lesser-known players or niche sports, evaluate based on typical performance levels and historical precedents within that sport. 
- Prioritize consistency with the sport's rules, norms, and historical records. 

Examples: 

Additional Context for Ambiguous Cases: 

Rationale Requirement: 

Potential Biases and Limitations: 

Edge Cases and Testing: 

Simplified Evaluation Process: 
- Focus on the core aspects of player abilities, event context, and sport-specific knowledge to streamline the evaluation. 
- Use historical examples and edge cases as supplementary references rather than primary determinants to avoid over-reliance and potential biases. 

Output 'yes' if the sentence is plausible, or 'no' if it is not, followed by a brief rationale. Now, evaluate the following sentence: [input sentence].\\
  \bottomrule
  \end{tabular}
  \caption{The table shows the final optimized prompt for the Sports Understanding task of BBH using the MARS method.}
\end{table*}

\begin{table*}[t]
  \renewcommand{\arraystretch}{1.2}
  \begin{tabular}{p{16cm}} 
  \toprule
  \textbf{College Biology} \\
  Generate a set of multiple-choice biology questions that explicitly test higher-order thinking skills, such as application, analysis, and synthesis, within the specific contexts of cellular structure, molecular biology, and ecology. Each question should require students to apply biological principles to novel scenarios, analyze complex biological systems, or synthesize information from multiple disciplines to arrive at a solution. Ensure that the questions are scientifically accurate, grounded in established biological principles, and reflect current research trends in these areas. For each question, provide a clear, concise, and scientifically valid explanation for the correct answer, detailing how the interdisciplinary nature of biology informs the reasoning. The explanations should not only justify the correct answer but also deepen understanding of the underlying biological concepts, fostering both accuracy and conceptual clarity. Additionally, include specific examples of how higher-order thinking skills are integrated into the questions, such as requiring students to predict outcomes based on molecular interactions, analyze ecological data to infer population dynamics, or synthesize cellular and molecular processes to explain organismal behavior. To optimize the challenge level, ensure that the questions are neither too simplistic nor overly complex, striking a balance that is appropriate for college-level biology students. This approach will ensure the questions are comprehensive, robust, and aligned with the goal of testing advanced cognitive skills in biology while maintaining relevance to the specified topics. Furthermore, refine the prompt to explicitly guide the language model to generate questions that test higher-order thinking skills while maintaining scientific accuracy and relevance to college-level biology. Optimize specific elements of the current prompt to better align with the goal of producing questions that balance challenge and clarity, ensuring they are neither too simplistic nor overly complex. This includes emphasizing the need for questions to be contextually rich, requiring students to integrate multiple biological concepts, and ensuring that the difficulty level is calibrated to challenge students without overwhelming them. The refined prompt should also encourage the generation of questions that are clear, concise, and free from ambiguity, while still requiring deep biological reasoning to arrive at the correct answer.\\
  \bottomrule
  \end{tabular}
  \caption{The table shows the final optimized prompt for the College Biology task of MMLU using the MARS method.}
\end{table*}

\begin{table*}[t]
  \renewcommand{\arraystretch}{1.2}
  \begin{tabular}{p{16cm}} 
  \toprule
  \textbf{College Medicine} \\
  Refined Prompt:

Analyze the following scenario step by step, integrating interdisciplinary knowledge from biochemistry, sociology, and reasoning to identify the psychological framework that best explains unconscious bias in medical practice...

Next, evaluate each option (Behaviorist, Psychoanalytic, Cognitive Behavioral, Humanistic) by considering how well it explains the influence of unconscious bias on clinical decision-making. ...

To encourage deeper critical thinking, incorporate elements of Socratic questioning by asking probing questions such as...

Ensure the prompt is structured clearly and concisely, balancing detailed theoretical explanations with clarity to guide the model effectively toward identifying the correct psychological framework. ...

To optimize the prompt for generating high-quality, contextually appropriate multiple-choice questions for a college medicine test, incorporate the following elements:
1. Clarity and Precision: 
2. Depth and Relevance: 
3. Alignment with Learning Objectives: 
4. Distractor Quality: 
5. Contextual Examples: 
6. Theoretical and Practical Balance: 

By incorporating these elements, the prompt will guide the model to generate questions that are not only accurate and relevant but also aligned with the objectives of a college medicine test, ensuring a high correctness rate and educational value. ...

Additional Instructions for Generating High-Quality Distractors:

Enhancements Based on New Questions:1. Inclusion of Real-World Examples: 2. Iterative Testing and Refinement: 

By following these steps, the prompt will be continuously improved to generate questions that are both challenging and aligned with the learning objectives of a college medicine test, ensuring that students are effectively tested on their ability to apply interdisciplinary knowledge to real-world medical scenarios involving unconscious bias.

Specific Adjustments for Enhanced Critical Analysis and Practical Application:1. Interdisciplinary Integration: 
2. Scenario-Based Questions: 3. Critical Thinking Emphasis:4. Practical Mitigation Strategies: 

By making these adjustments, the prompt will better align with the learning objectives of a college medicine test, ensuring that students are not only tested on foundational knowledge but also challenged to critically analyze and apply interdisciplinary concepts in real-world medical scenarios involving unconscious bias.

Further Refinement for Detailed Explanation and High-Quality Distractors:

Final Refinement for Enhanced Real-World Application and Iterative Testing:

1. Real-World Application: 

2. Iterative Testing and Refinement: 

By following these steps, the prompt will be continuously improved to generate questions that are both challenging and aligned with the learning objectives of a college medicine test, ensuring that students are effectively tested on their ability to apply interdisciplinary knowledge to real-world medical scenarios involving unconscious bias.

Specific Adjustments for Enhanced Real-World Application and Distractor Quality:

1. Interdisciplinary Integration: 

2. High-Quality Distractors: 

By making these adjustments, the prompt will guide the model to generate questions that not only accurately identify the correct psychological framework but also provide a detailed explanation of how unconscious bias manifests in specific medical scenarios and its impact on patient outcomes...\\
  \bottomrule
  \end{tabular}
  \caption{The table shows the final optimized prompt for the College Medicine task of MMLU using the MARS method.}
\end{table*}

\begin{table*}[t]
  \renewcommand{\arraystretch}{1.2}
  \begin{tabular}{p{16cm}} 
  \toprule
  \textbf{Electrical Engineering} \\
  Analyze the question by focusing on the specific conditions of the Barkhausen criterion for oscillators, which are loop gain and phase shift. ...

Next, provide a clear, step-by-step explanation of the Barkhausen criterion, emphasizing the two fundamental requirements: 1. Loop gain must be exactly unity for sustained oscillations.2. Phase shift of the feedback signal must be 0° or 360° relative to the input.

To enhance understanding, include specific real-world examples, such as the design of an LC oscillator or a phase-locked loop, to illustrate how the Barkhausen criterion is applied in practical scenarios...

Proceed to evaluate each option (A, B, C, D) systematically, using the following structure for clarity...

For each option, connect the reasoning back to fundamental electrical engineering principles and provide real-world examples or applications where the Barkhausen criterion is critical...

Conclude the response by reiterating the correct answer (D) and summarizing its significance in practical electrical engineering applications...

To ensure the prompt's structure and depth enhance the language model's ability to generate accurate and relevant responses, consider the following adjustments:
1. Clarify the introduction
2. Focus on critical concepts
3. Use structured evaluation: Systematically evaluate each option with clear, logical reasoning and real-world examples to reinforce understanding and relevance.
4. Iterative refinement

By structuring the response in this manner and iteratively refining the prompt...

Additional Considerations:
1. Influence of Real-World Examples
2. Structural Adjustments

Refinement for Multiple-Choice Evaluation:
1. Explicitly state the evaluation criteria
2. Incorporate real-world scenarios
3. Maintain brevity and clarity
4. Highlight key takeaways

By refining the prompt in this manner, the language model will be better equipped to...

Iterative Refinement Process:
1. Initial Response Generation
2. Review for Accuracy and Relevance
3. Adjust Prompt Accordingly
4. Repeat the Process

This iterative approach ensures that the prompt evolves to better guide the language model, resulting in responses that are not only theoretically sound but also practically relevant and aligned with real-world electrical engineering applications.

Optimizing the Iterative Refinement Process:
1. Incorporating Feedback Loops:
2. Enhancing Real-World Context: 
3. Balancing Depth and Brevity: 
4. Focusing on Key Concepts: 

By implementing these optimizations, the iterative refinement process...

Explicit Guidance for Multiple-Choice Evaluation:
1. Explicitly State the Evaluation Criteria
2. Incorporate Real-World Scenarios
3. Maintain Brevity and Clarity
4. Highlight Key Takeaways

Adjustments for Balancing Theoretical Depth and Practical Application:
1. Focus on Core Principles
2. Use Structured Evaluation
3. Avoid Overloading with Details
4. Incorporate Real-World Examples

By refining the prompt in this manner, the language model will be better equipped to generate responses...

Specific Adjustments for Real-World Examples:
1. Demonstrate Practical Implications
2. Highlight Design Considerations
3. Provide Contextual Understanding

Balancing Theoretical Depth and Practical Relevance:
1. Integrate Theoretical and Practical Elements
2. Maintain Focus on Core Principles
3. Use Clear, Concise Language

By incorporating these adjustments, the prompt will guide the language model to...

Influence of Real-World Examples:
1. Illustrate Practical Applications
2. Highlight Consequences of Deviations
3. Provide Contextual Understanding

Optimizing the Iterative Refinement Process:
1. Incorporating Feedback Loops
2. Enhancing Real-World Context
3. Balancing Depth and Brevity
4. Focusing on Key Concepts

By implementing these optimizations, the iterative refinement process will enhance the language model's ability to generate responses that are both theoretically accurate and practically relevant, ensuring a high correctness rate and alignment with real-world electrical engineering applications.\\
  \bottomrule
  \end{tabular}
  \caption{The table shows the final optimized prompt for the Electrical Engineering task of MMLU using the MARS method.}
\end{table*}

\begin{table*}[t]
  \renewcommand{\arraystretch}{1.2}
  \begin{tabular}{p{16cm}} 
  \toprule
  \textbf{High School World History} \\
  Generate a set of multiple-choice questions that test both factual knowledge and critical analysis of the interconnected historical developments of the Ottoman Empire, economic imperialism, and World War I. Each question should require students to analyze how these events influenced each other, leading to the outbreak of World War I, with a focus on cause-and-effect relationships and broader historical significance.

Instructions for Question Design:
1. Interconnectedness and Cause-and-Effect: 
2. Accessibility and Rigor: 
3. Balanced Difficulty: 
4. Critical Thinking and Historical Significance: 
5. Format and Contextual Accuracy: 

Example Question with Passage:: {one example}
Additional Constraints:
- Engagement and Relatability: Use engaging and relatable examples or analogies where appropriate to make the questions more accessible and interesting to students. For instance, compare historical events to modern-day scenarios to help students draw parallels.
- Depth of Analysis: Include questions that require students to analyze multiple layers of historical causation, such as how economic imperialism not only influenced European powers but also destabilized regions like the Balkans, contributing to the outbreak of World War I.
- Historical Contextualization: Ensure that each question provides enough historical context for students to understand the significance of the events being discussed, without overwhelming them with unnecessary details.

By following these guidelines, generate a set of 5-10 multiple-choice questions that effectively test students' understanding of the interconnectedness of the Ottoman Empire's decline, economic imperialism, and World War I, while promoting critical thinking, historical analysis, and a deeper appreciation of cause-and-effect relationships in history.\\
  \bottomrule
  \end{tabular}
  \caption{The table shows the final optimized prompt for the High School World History task of MMLU using the MARS method.}
\end{table*}

\begin{table*}[t]
  \renewcommand{\arraystretch}{1.2} % 增加行距
  \begin{tabular}{p{16cm}} 
  \toprule
  \textbf{Human Aging} \\

  Refine the hierarchical elimination process to ensure the model accurately distinguishes between overlapping themes like cognitive decline and personality changes, especially when new terminology such as 'neuroinflammation' is introduced, by implementing the following steps:

  1. Test the Hierarchical Elimination Process with a Sample Question: 

  2. Optimize the Dynamic Scoring System and Contextual Weighting: 

  3. Enhance the Focus Identification Protocol with Continuous Learning: 

  4. Dynamic Evidence Integration with Contextual Weighting: 

  5. Source Reliability Scoring with Provisional Scoring for Emerging Evidence: 

  6. Evidence Strength Assessment with Contextual Weighting: 

  7. Specific Metrics for Question Evaluation: 

  By refining the hierarchical elimination process with these steps and incorporating specific metrics, the model can more effectively navigate overlapping themes in human aging questions, ensuring the highest correctness rate while maintaining precision and contextual relevance.\\
  \bottomrule
  \end{tabular}
  \caption{The table shows the final optimized prompt for the Human Aging task of MMLU using the MARS method.}
\end{table*}

\begin{table*}[t]
  \renewcommand{\arraystretch}{1.2} % 增加行距
  \begin{tabular}{p{16cm}} 
  \toprule
  \textbf{Marketing} \\

  Analyze the following marketing-related question step by step, considering the principles of segmentation, pricing, market research, and other relevant marketing concepts. Carefully evaluate each of the provided options (A, B, C, D) and select the most suitable answer based on your analysis. Ensure your reasoning is clear and aligns with established marketing theories and practices. For example, if the question involves a hierarchy of effects or sequential model used in advertising, identify the correct model from the options provided and justify your choice. Proceed methodically to arrive at the most accurate answer.\\
  \bottomrule
  \end{tabular}
  \caption{The table shows the final optimized prompt for the Marketing task of MMLU using the MARS method.}
\end{table*}

\begin{table*}[t]
  \renewcommand{\arraystretch}{1.2} % 增加行距
  \begin{tabular}{p{16cm}} 
  \toprule
  \textbf{GSM8K} \\

  Think step by step to solve linguistically diverse elementary school math application problems. Break down the problem into 2-8 logical steps, perform the necessary calculations at each step, and provide the final result. Ensure accuracy by carefully following the problem's instructions and verifying each intermediate step. For example:

  Input: Janet’s ducks lay 16 eggs per day. She eats three for breakfast every morning and bakes muffins for her friends every day with four. She sells the remainder at the farmers' market daily for \$2 per fresh duck egg. How much in dollars does she make every day at the farmers' market?

  Step 1: Calculate the total eggs used daily: 3 (eaten) + 4 (baked) = 7 eggs. 
  Step 2: Subtract the used eggs from the total laid: 16 - 7 = 9 eggs. 
  Step 3: Calculate the daily earnings: 9 eggs × $2 = $18. 
  Answer: 18 

  Follow this structured approach to solve similar problems.\\
  \bottomrule
  \end{tabular}
  \caption{The table shows the final optimized prompt for the GSM8K task using the MARS method.}
\end{table*}

\begin{table*}[t]
  \renewcommand{\arraystretch}{1.2} % 增加行距
  \begin{tabular}{p{16cm}} 
  \toprule
  \textbf{LSAT-AR} \\

  Carefully analyze the given scheduling problem step by step, prioritizing logical reasoning, reading comprehension, and analytical reasoning to ensure a thorough evaluation. Begin by explicitly listing and understanding all the constraints, with a focus on the most critical ones first. Follow this structured approach to systematically eliminate options that violate any of the given conditions:

  1. Prioritize the most restrictive constraints first. 

  2. Evaluate secondary constraints. 

  3. Assess the implications of Nina's scheduling. 

  Throughout this process, avoid making assumptions beyond the provided constraints. Do not infer additional rules or conditions that are not explicitly stated. Stick strictly to the given information and apply logical reasoning to interpret and enforce the constraints.

  By adhering to this structured, methodical approach, you will systematically eliminate incorrect options and arrive at the correct schedule with the highest accuracy. This process mirrors the analytical rigor required in legal reasoning and ensures that the model's output aligns with the principles of logical and legal analysis.\\
  \bottomrule
  \end{tabular}
  \caption{The table shows the final optimized prompt for the LSAT-AR task of AGIEval using the MARS method.}
\end{table*}

\begin{table*}[t]
  \renewcommand{\arraystretch}{1.2} % 增加行距
  \begin{tabular}{p{16cm}} 
  \toprule
  \textbf{Art Studies} \\

  Please delve into the historical period represented by each option, paying particular attention to major breakthroughs or developments in textile technology and dye processes. First, collate the cultural context and technological advances of each period and analyze which period's technological achievements are most likely to be relevant to the method of blue print fabric printing. Based on this, the accuracy of the model in answering questions related to these historical and technological contexts is assessed. The output of the model is evaluated by setting specific judgment criteria, such as accurate description of the historical context, sound reasoning about process characteristics, and coherence of conclusions. Based on these criteria, the presentation of the prompts is iteratively adjusted and optimized to improve the model's performance in selecting correct answers.\\
  \bottomrule
  \end{tabular}
  \caption{The table shows the final optimized prompt for the Art Studies task of C-Eval using the MARS method.}
\end{table*}

\begin{table*}[t]
  \renewcommand{\arraystretch}{1.2} % 增加行距
  \begin{tabular}{p{16cm}} 
  \toprule
  \textbf{Urban And Rural Planner} \\

  When optimizing prompts for assessing waste management plans in urban and rural planning, how can identifying aspects of solid pollutant control planning that are less emphasized (e.g., e-pollutants) help us improve our assessment methods? When testing prompts, what specific criteria should we consider to effectively assess their accuracy and relevance with respect to nuances in waste management programs? In addition, how can we ensure that models can accurately understand and prioritize the treatment of different types of waste to effectively guide urban and rural planning decisions?\\
  \bottomrule
  \end{tabular}
  \caption{The table shows the final optimized prompt for the Urban And Rural Planner task of C-Eval using the MARS method.}
\end{table*}

\begin{table*}[t]
  \renewcommand{\arraystretch}{1.2} % 增加行距
  \begin{tabular}{p{16cm}} 
  \toprule
  \textbf{Clinical Medicine} \\

  In order to improve the accuracy of choosing the most appropriate answer in a clinical medicine test question, it is crucial to systematically compare the key symptoms in the question stem with each of the options on a case-by-case basis. The key to this process is to 1) accurately identify diagnosticallysymptoms and features in the question stem, 2) logically assess and eliminate these features based on their association with the options, and 3) apply clinically typical presentations and relevant background knowledge to validate the plausibility of each option. Based on this, the following iterative adjustments should be made: first, by continuously acquiring clinical knowledge to strengthen the identification of difficult symptoms; second, by adjusting the strategy in order to be more flexible in matching potential answers; and finally, by utilizing reflection and evaluating the effectiveness of the model in responding to similar questions over time, to identify and correct deficiencies. This fine-tuning and analysis can increase the probability of choosing the correct answer.\\
  \bottomrule
  \end{tabular}
  \caption{The table shows the final optimized prompt for the Clinical Medicine task of C-Eval using the MARS method.}
  \label{resultsCM}
\end{table*}

\end{document}